\documentclass[accepted]{uai2022} % after acceptance, for a revised
\usepackage[american]{babel}
\usepackage{natbib} % has a nice set of citation styles and commands
\usepackage{mathtools} % amsmath with fixes and additions
\usepackage{booktabs} % commands to create good-looking tables
\usepackage{tikz} % nice language for creating drawings and diagrams

\usepackage[FIGBOTCAP,TABBOTCAP,center,nooneline]{subfigure}
\usepackage[dvips]{rotating}
\usepackage[noend]{RSalgorithmic}
\usepackage{algorithm}
\usepackage{amscd}
\usepackage{amsfonts}
\usepackage{amssymb}
\usepackage{color}
\usepackage{comment}
\usepackage{datetime}
\usepackage{epsfig}
\usepackage{xifthen}
\usepackage{latexsym}
\usepackage{marginnote}
\usepackage{multirow}
\usepackage{placeins}
\usepackage{rotate} 
\usepackage{stmaryrd}
\usepackage{tikz}
\usepackage[draft]{todonotes}
\usepackage{varwidth}
\usepackage{verbatim} 
\usepackage{wrapfig}
\usepackage{xcolor,colortbl}
\usepackage{xspace}

\usepackage[dvips]{rotating}
\usepackage[noend]{RSalgorithmic}
\usepackage{algorithm}
\usepackage{amscd}
\usepackage{amsfonts}
\usepackage{amssymb}
\usepackage{color}
\usepackage{comment}
\usepackage{datetime}
\usepackage{epsfig}
\usepackage{ifthen}
\usepackage{latexsym}
\usepackage{marginnote}
\usepackage{multirow}
\usepackage{placeins}
\usepackage{rotate} 
\usepackage{stmaryrd}
\usepackage{tikz}
\usepackage{varwidth}
\usepackage{verbatim} 
\usepackage{wrapfig}
\usepackage{xcolor,colortbl}
\usepackage{xspace}
\usepackage{caption}
\usepackage{colour-blind}
\usepackage{lipsum}  

\usepackage[dvips]{rotating}
\usepackage[noend]{RSalgorithmic}
\usepackage{algorithm}
\usepackage{amscd}
\usepackage{amsfonts}
\usepackage{amssymb}
\usepackage{color}
\usepackage{comment}
\usepackage{datetime}
\usepackage{epsfig}
\usepackage{xifthen}
\usepackage{latexsym}
\usepackage{marginnote}
\usepackage{multirow}
\usepackage{placeins}
\usepackage{rotate} 
\usepackage{stmaryrd}
\usepackage{tikz}
\usepackage[draft]{todonotes}
\usepackage{varwidth}
\usepackage{verbatim} 
\usepackage{wrapfig}
\usepackage{xcolor,colortbl}
\usepackage{xspace}

\usepackage[dvips]{rotating}
\usepackage[noend]{RSalgorithmic}
\usepackage{algorithm}
\usepackage{amscd}
\usepackage{amsfonts}
\usepackage{amssymb}
\usepackage{color}
\usepackage{comment}
\usepackage{datetime}
\usepackage{epsfig}
\usepackage{ifthen}
\usepackage{latexsym}
\usepackage{marginnote}
\usepackage{multirow}
\usepackage{placeins}
\usepackage{rotate} 
\usepackage{stmaryrd}
\usepackage{tikz}
\usepackage{varwidth}
\usepackage{verbatim} 
\usepackage{wrapfig}
\usepackage{xcolor,colortbl}
\usepackage{xspace}
\usepackage{caption}
\usepackage{colour-blind}
\usepackage{lipsum}  
\newtheorem{example}{Example}

\definecolor {snow}                {rgb}{1.00,0.98,0.98}
\definecolor {ghostwhite}          {rgb}{0.97,0.97,1.00}
\definecolor {whitesmoke}          {rgb}{0.96,0.96,0.96}
\definecolor {gainsboro}           {rgb}{0.86,0.86,0.86}
\definecolor {floralwhite}         {rgb}{1.00,0.98,0.94}
\definecolor {oldlace}             {rgb}{0.99,0.96,0.90}
\definecolor {linen}               {rgb}{0.98,0.94,0.90}
\definecolor {antiquewhite}        {rgb}{0.98,0.92,0.84}
\definecolor {papayawhip}          {rgb}{1.00,0.94,0.84}
\definecolor {blanchedalmond}      {rgb}{1.00,0.92,0.80}
\definecolor {bisque}              {rgb}{1.00,0.89,0.77}
\definecolor {peachpuff}           {rgb}{1.00,0.85,0.73}
\definecolor {navajowhite}         {rgb}{1.00,0.87,0.68}
\definecolor {moccasin}            {rgb}{1.00,0.89,0.71}
\definecolor {cornsilk}            {rgb}{1.00,0.97,0.86}
\definecolor {ivory}               {rgb}{1.00,1.00,0.94}
\definecolor {lemonchiffon}        {rgb}{1.00,0.98,0.80}
\definecolor {seashell}            {rgb}{1.00,0.96,0.93}
\definecolor {honeydew}            {rgb}{0.94,1.00,0.94}
\definecolor {mintcream}           {rgb}{0.96,1.00,0.98}
\definecolor {azure}               {rgb}{0.94,1.00,1.00}
\definecolor {aliceblue}           {rgb}{0.94,0.97,1.00}
\definecolor {lavender}            {rgb}{0.90,0.90,0.98}
\definecolor {lavenderblush}       {rgb}{1.00,0.94,0.96}
\definecolor {mistyrose}           {rgb}{1.00,0.89,0.88}
\definecolor {white}               {rgb}{1.00,1.00,1.00}
\definecolor {black}               {rgb}{0.00,0.00,0.00}
\definecolor {darkslategray}       {rgb}{0.18,0.31,0.31}
\definecolor {dimgray}             {rgb}{0.41,0.41,0.41}
\definecolor {slategray}           {rgb}{0.44,0.50,0.56}
\definecolor {lightslategray}      {rgb}{0.47,0.53,0.60}
\definecolor {gray}                {rgb}{0.75,0.75,0.75}
\definecolor {lightgrey}           {rgb}{0.83,0.83,0.83}
\definecolor {midnightblue}        {rgb}{0.10,0.10,0.44}
\definecolor {navy}                {rgb}{0.00,0.00,0.50}
\definecolor {cornflowerblue}      {rgb}{0.39,0.58,0.93}
\definecolor {darkslateblue}       {rgb}{0.28,0.24,0.55}
\definecolor {slateblue}           {rgb}{0.42,0.35,0.80}
\definecolor {mediumslateblue}     {rgb}{0.48,0.41,0.93}
\definecolor {lightslateblue}      {rgb}{0.52,0.44,1.00}
\definecolor {mediumblue}          {rgb}{0.00,0.00,0.80}
\definecolor {royalblue}           {rgb}{0.25,0.41,0.88}
\definecolor {blue}                {rgb}{0.00,0.00,1.00}
\definecolor {dodgerblue}          {rgb}{0.12,0.56,1.00}
\definecolor {deepskyblue}         {rgb}{0.00,0.75,1.00}
\definecolor {skyblue}             {rgb}{0.53,0.81,0.92}
\definecolor {lightskyblue}        {rgb}{0.53,0.81,0.98}
\definecolor {steelblue}           {rgb}{0.27,0.51,0.71}
\definecolor {lightsteelblue}      {rgb}{0.69,0.77,0.87}
\definecolor {lightblue}           {rgb}{0.68,0.85,0.90}
\definecolor {powderblue}          {rgb}{0.69,0.88,0.90}
\definecolor {paleturquoise}       {rgb}{0.69,0.93,0.93}
\definecolor {darkturquoise}       {rgb}{0.00,0.81,0.82}
\definecolor {mediumturquoise}     {rgb}{0.28,0.82,0.80}
\definecolor {turquoise}           {rgb}{0.25,0.88,0.82}
\definecolor {cyan}                {rgb}{0.00,1.00,1.00}
\definecolor {lightcyan}           {rgb}{0.88,1.00,1.00}
\definecolor {cadetblue}           {rgb}{0.37,0.62,0.63}
\definecolor {mediumaquamarine}    {rgb}{0.40,0.80,0.67}
\definecolor {aquamarine}          {rgb}{0.50,1.00,0.83}
\definecolor {darkgreen}           {rgb}{0.00,0.39,0.00}
\definecolor {darkolivegreen}      {rgb}{0.33,0.42,0.18}
\definecolor {darkseagreen}        {rgb}{0.56,0.74,0.56}
\definecolor {seagreen}            {rgb}{0.18,0.55,0.34}
\definecolor {mediumseagreen}      {rgb}{0.24,0.70,0.44}
\definecolor {lightseagreen}       {rgb}{0.13,0.70,0.67}
\definecolor {palegreen}           {rgb}{0.60,0.98,0.60}
\definecolor {springgreen}         {rgb}{0.00,1.00,0.50}
\definecolor {lawngreen}           {rgb}{0.49,0.99,0.00}
\definecolor {green}               {rgb}{0.00,1.00,0.00}
\definecolor {chartreuse}          {rgb}{0.50,1.00,0.00}
\definecolor {mediumspringgreen}   {rgb}{0.00,0.98,0.60}
\definecolor {greenyellow}         {rgb}{0.68,1.00,0.18}
\definecolor {limegreen}           {rgb}{0.20,0.80,0.20}
\definecolor {yellowgreen}         {rgb}{0.60,0.80,0.20}
\definecolor {forestgreen}         {rgb}{0.13,0.55,0.13}
\definecolor {olivedrab}           {rgb}{0.42,0.56,0.14}
\definecolor {darkkhaki}           {rgb}{0.74,0.72,0.42}
\definecolor {khaki}               {rgb}{0.94,0.90,0.55}
\definecolor {palegoldenrod}       {rgb}{0.93,0.91,0.67}
\definecolor {lightgoldenrodyellow} {rgb}{0.98,0.98,0.82}
\definecolor {lightyellow}         {rgb}{1.00,1.00,0.88}
\definecolor {yellow}              {rgb}{1.00,1.00,0.00}
\definecolor {gold}                {rgb}{1.00,0.84,0.00}
\definecolor {lightgoldenrod}      {rgb}{0.93,0.87,0.51}
\definecolor {goldenrod}           {rgb}{0.85,0.65,0.13}
\definecolor {darkgoldenrod}       {rgb}{0.72,0.53,0.04}
\definecolor {rosybrown}           {rgb}{0.74,0.56,0.56}
\definecolor {indianred}           {rgb}{0.80,0.36,0.36}
\definecolor {saddlebrown}         {rgb}{0.55,0.27,0.07}
\definecolor {sienna}              {rgb}{0.63,0.32,0.18}
\definecolor {peru}                {rgb}{0.80,0.52,0.25}
\definecolor {burlywood}           {rgb}{0.87,0.72,0.53}
\definecolor {beige}               {rgb}{0.96,0.96,0.86}
\definecolor {wheat}               {rgb}{0.96,0.87,0.70}
\definecolor {sandybrown}          {rgb}{0.96,0.64,0.38}
\definecolor {tan}                 {rgb}{0.82,0.71,0.55}
\definecolor {chocolate}           {rgb}{0.82,0.41,0.12}
\definecolor {firebrick}           {rgb}{0.70,0.13,0.13}
\definecolor {brown}               {rgb}{0.65,0.16,0.16}
\definecolor {darksalmon}          {rgb}{0.91,0.59,0.48}
\definecolor {salmon}              {rgb}{0.98,0.50,0.45}
\definecolor {lightsalmon}         {rgb}{1.00,0.63,0.48}
\definecolor {orange}              {rgb}{1.00,0.65,0.00}
\definecolor {darkorange}          {rgb}{1.00,0.55,0.00}
\definecolor {coral}               {rgb}{1.00,0.50,0.31}
\definecolor {lightcoral}          {rgb}{0.94,0.50,0.50}
\definecolor {tomato}              {rgb}{1.00,0.39,0.28}
\definecolor {orangered}           {rgb}{1.00,0.27,0.00}
\definecolor {red}                 {rgb}{1.00,0.00,0.00}
\definecolor {hotpink}             {rgb}{1.00,0.41,0.71}
\definecolor {deeppink}            {rgb}{1.00,0.08,0.58}
\definecolor {pink}                {rgb}{1.00,0.75,0.80}
\definecolor {lightpink}           {rgb}{1.00,0.71,0.76}
\definecolor {palevioletred}       {rgb}{0.86,0.44,0.58}
\definecolor {maroon}              {rgb}{0.69,0.19,0.38}
\definecolor {mediumvioletred}     {rgb}{0.78,0.08,0.52}
\definecolor {violetred}           {rgb}{0.82,0.13,0.56}
\definecolor {magenta}             {rgb}{1.00,0.00,1.00}
\definecolor {violet}              {rgb}{0.93,0.51,0.93}
\definecolor {plum}                {rgb}{0.87,0.63,0.87}
\definecolor {orchid}              {rgb}{0.85,0.44,0.84}
\definecolor {mediumorchid}        {rgb}{0.73,0.33,0.83}
\definecolor {darkorchid}          {rgb}{0.60,0.20,0.80}
\definecolor {darkviolet}          {rgb}{0.58,0.00,0.83}
\definecolor {blueviolet}          {rgb}{0.54,0.17,0.89}
\definecolor {purple}              {rgb}{0.63,0.13,0.94}
\definecolor {mediumpurple}        {rgb}{0.58,0.44,0.86}
\definecolor {thistle}             {rgb}{0.85,0.75,0.85}
\definecolor {snow2}               {rgb}{0.93,0.91,0.91}
\definecolor {snow3}               {rgb}{0.80,0.79,0.79}
\definecolor {snow4}               {rgb}{0.55,0.54,0.54}
\definecolor {seashell2}           {rgb}{0.93,0.90,0.87}
\definecolor {seashell3}           {rgb}{0.80,0.77,0.75}
\definecolor {seashell4}           {rgb}{0.55,0.53,0.51}
\definecolor {antiquewhite1}       {rgb}{1.00,0.94,0.86}
\definecolor {antiquewhite2}       {rgb}{0.93,0.87,0.80}
\definecolor {antiquewhite3}       {rgb}{0.80,0.75,0.69}
\definecolor {antiquewhite4}       {rgb}{0.55,0.51,0.47}
\definecolor {bisque2}             {rgb}{0.93,0.84,0.72}
\definecolor {bisque3}             {rgb}{0.80,0.72,0.62}
\definecolor {bisque4}             {rgb}{0.55,0.49,0.42}
\definecolor {peachpuff2}          {rgb}{0.93,0.80,0.68}
\definecolor {peachpuff3}          {rgb}{0.80,0.69,0.58}
\definecolor {peachpuff4}          {rgb}{0.55,0.47,0.40}
\definecolor {navajowhite2}        {rgb}{0.93,0.81,0.63}
\definecolor {navajowhite3}        {rgb}{0.80,0.70,0.55}
\definecolor {navajowhite4}        {rgb}{0.55,0.47,0.37}
\definecolor {lemonchiffon2}       {rgb}{0.93,0.91,0.75}
\definecolor {lemonchiffon3}       {rgb}{0.80,0.79,0.65}
\definecolor {lemonchiffon4}       {rgb}{0.55,0.54,0.44}
\definecolor {cornsilk2}           {rgb}{0.93,0.91,0.80}
\definecolor {cornsilk3}           {rgb}{0.80,0.78,0.69}
\definecolor {cornsilk4}           {rgb}{0.55,0.53,0.47}
\definecolor {ivory2}              {rgb}{0.93,0.93,0.88}
\definecolor {ivory3}              {rgb}{0.80,0.80,0.76}
\definecolor {ivory4}              {rgb}{0.55,0.55,0.51}
\definecolor {honeydew2}           {rgb}{0.88,0.93,0.88}
\definecolor {honeydew3}           {rgb}{0.76,0.80,0.76}
\definecolor {honeydew4}           {rgb}{0.51,0.55,0.51}
\definecolor {lavenderblush2}      {rgb}{0.93,0.88,0.90}
\definecolor {lavenderblush3}      {rgb}{0.80,0.76,0.77}
\definecolor {lavenderblush4}      {rgb}{0.55,0.51,0.53}
\definecolor {mistyrose2}          {rgb}{0.93,0.84,0.82}
\definecolor {mistyrose3}          {rgb}{0.80,0.72,0.71}
\definecolor {mistyrose4}          {rgb}{0.55,0.49,0.48}
\definecolor {azure2}              {rgb}{0.88,0.93,0.93}
\definecolor {azure3}              {rgb}{0.76,0.80,0.80}
\definecolor {azure4}              {rgb}{0.51,0.55,0.55}
\definecolor {slateblue1}          {rgb}{0.51,0.44,1.00}
\definecolor {slateblue2}          {rgb}{0.48,0.40,0.93}
\definecolor {slateblue3}          {rgb}{0.41,0.35,0.80}
\definecolor {slateblue4}          {rgb}{0.28,0.24,0.55}
\definecolor {royalblue1}          {rgb}{0.28,0.46,1.00}
\definecolor {royalblue2}          {rgb}{0.26,0.43,0.93}
\definecolor {royalblue3}          {rgb}{0.23,0.37,0.80}
\definecolor {royalblue4}          {rgb}{0.15,0.25,0.55}
\definecolor {blue2}               {rgb}{0.00,0.00,0.93}
\definecolor {blue4}               {rgb}{0.00,0.00,0.55}
\definecolor {dodgerblue2}         {rgb}{0.11,0.53,0.93}
\definecolor {dodgerblue3}         {rgb}{0.09,0.45,0.80}
\definecolor {dodgerblue4}         {rgb}{0.06,0.31,0.55}
\definecolor {steelblue1}          {rgb}{0.39,0.72,1.00}
\definecolor {steelblue2}          {rgb}{0.36,0.67,0.93}
\definecolor {steelblue3}          {rgb}{0.31,0.58,0.80}
\definecolor {steelblue4}          {rgb}{0.21,0.39,0.55}
\definecolor {deepskyblue2}        {rgb}{0.00,0.70,0.93}
\definecolor {deepskyblue3}        {rgb}{0.00,0.60,0.80}
\definecolor {deepskyblue4}        {rgb}{0.00,0.41,0.55}
\definecolor {skyblue1}            {rgb}{0.53,0.81,1.00}
\definecolor {skyblue2}            {rgb}{0.49,0.75,0.93}
\definecolor {skyblue3}            {rgb}{0.42,0.65,0.80}
\definecolor {skyblue4}            {rgb}{0.29,0.44,0.55}
\definecolor {lightskyblue1}       {rgb}{0.69,0.89,1.00}
\definecolor {lightskyblue2}       {rgb}{0.64,0.83,0.93}
\definecolor {lightskyblue3}       {rgb}{0.55,0.71,0.80}
\definecolor {lightskyblue4}       {rgb}{0.38,0.48,0.55}
\definecolor {slategray1}          {rgb}{0.78,0.89,1.00}
\definecolor {slategray2}          {rgb}{0.73,0.83,0.93}
\definecolor {slategray3}          {rgb}{0.62,0.71,0.80}
\definecolor {slategray4}          {rgb}{0.42,0.48,0.55}
\definecolor {lightsteelblue1}     {rgb}{0.79,0.88,1.00}
\definecolor {lightsteelblue2}     {rgb}{0.74,0.82,0.93}
\definecolor {lightsteelblue3}     {rgb}{0.64,0.71,0.80}
\definecolor {lightsteelblue4}     {rgb}{0.43,0.48,0.55}
\definecolor {lightblue1}          {rgb}{0.75,0.94,1.00}
\definecolor {lightblue2}          {rgb}{0.70,0.87,0.93}
\definecolor {lightblue3}          {rgb}{0.60,0.75,0.80}
\definecolor {lightblue4}          {rgb}{0.41,0.51,0.55}
\definecolor {lightcyan2}          {rgb}{0.82,0.93,0.93}
\definecolor {lightcyan3}          {rgb}{0.71,0.80,0.80}
\definecolor {lightcyan4}          {rgb}{0.48,0.55,0.55}
\definecolor {paleturquoise1}      {rgb}{0.73,1.00,1.00}
\definecolor {paleturquoise2}      {rgb}{0.68,0.93,0.93}
\definecolor {paleturquoise3}      {rgb}{0.59,0.80,0.80}
\definecolor {paleturquoise4}      {rgb}{0.40,0.55,0.55}
\definecolor {cadetblue1}          {rgb}{0.60,0.96,1.00}
\definecolor {cadetblue2}          {rgb}{0.56,0.90,0.93}
\definecolor {cadetblue3}          {rgb}{0.48,0.77,0.80}
\definecolor {cadetblue4}          {rgb}{0.33,0.53,0.55}
\definecolor {turquoise1}          {rgb}{0.00,0.96,1.00}
\definecolor {turquoise2}          {rgb}{0.00,0.90,0.93}
\definecolor {turquoise3}          {rgb}{0.00,0.77,0.80}
\definecolor {turquoise4}          {rgb}{0.00,0.53,0.55}
\definecolor {cyan2}               {rgb}{0.00,0.93,0.93}
\definecolor {cyan3}               {rgb}{0.00,0.80,0.80}
\definecolor {cyan4}               {rgb}{0.00,0.55,0.55}
\definecolor {darkslategray1}      {rgb}{0.59,1.00,1.00}
\definecolor {darkslategray2}      {rgb}{0.55,0.93,0.93}
\definecolor {darkslategray3}      {rgb}{0.47,0.80,0.80}
\definecolor {darkslategray4}      {rgb}{0.32,0.55,0.55}
\definecolor {aquamarine2}         {rgb}{0.46,0.93,0.78}
\definecolor {aquamarine4}         {rgb}{0.27,0.55,0.45}
\definecolor {darkseagreen1}       {rgb}{0.76,1.00,0.76}
\definecolor {darkseagreen2}       {rgb}{0.71,0.93,0.71}
\definecolor {darkseagreen3}       {rgb}{0.61,0.80,0.61}
\definecolor {darkseagreen4}       {rgb}{0.41,0.55,0.41}
\definecolor {seagreen1}           {rgb}{0.33,1.00,0.62}
\definecolor {seagreen2}           {rgb}{0.31,0.93,0.58}
\definecolor {seagreen3}           {rgb}{0.26,0.80,0.50}
\definecolor {palegreen1}          {rgb}{0.60,1.00,0.60}
\definecolor {palegreen2}          {rgb}{0.56,0.93,0.56}
\definecolor {palegreen3}          {rgb}{0.49,0.80,0.49}
\definecolor {palegreen4}          {rgb}{0.33,0.55,0.33}
\definecolor {springgreen2}        {rgb}{0.00,0.93,0.46}
\definecolor {springgreen3}        {rgb}{0.00,0.80,0.40}
\definecolor {springgreen4}        {rgb}{0.00,0.55,0.27}
\definecolor {green2}              {rgb}{0.00,0.93,0.00}
\definecolor {green3}              {rgb}{0.00,0.80,0.00}
\definecolor {green4}              {rgb}{0.00,0.55,0.00}
\definecolor {chartreuse2}         {rgb}{0.46,0.93,0.00}
\definecolor {chartreuse3}         {rgb}{0.40,0.80,0.00}
\definecolor {chartreuse4}         {rgb}{0.27,0.55,0.00}
\definecolor {olivedrab1}          {rgb}{0.75,1.00,0.24}
\definecolor {olivedrab2}          {rgb}{0.70,0.93,0.23}
\definecolor {olivedrab4}          {rgb}{0.41,0.55,0.13}
\definecolor {darkolivegreen1}     {rgb}{0.79,1.00,0.44}
\definecolor {darkolivegreen2}     {rgb}{0.74,0.93,0.41}
\definecolor {darkolivegreen3}     {rgb}{0.64,0.80,0.35}
\definecolor {darkolivegreen4}     {rgb}{0.43,0.55,0.24}
\definecolor {khaki1}              {rgb}{1.00,0.96,0.56}
\definecolor {khaki2}              {rgb}{0.93,0.90,0.52}
\definecolor {khaki3}              {rgb}{0.80,0.78,0.45}
\definecolor {khaki4}              {rgb}{0.55,0.53,0.31}
\definecolor {lightgoldenrod1}     {rgb}{1.00,0.93,0.55}
\definecolor {lightgoldenrod2}     {rgb}{0.93,0.86,0.51}
\definecolor {lightgoldenrod3}     {rgb}{0.80,0.75,0.44}
\definecolor {lightgoldenrod4}     {rgb}{0.55,0.51,0.30}
\definecolor {lightyellow2}        {rgb}{0.93,0.93,0.82}
\definecolor {lightyellow3}        {rgb}{0.80,0.80,0.71}
\definecolor {lightyellow4}        {rgb}{0.55,0.55,0.48}
\definecolor {yellow2}             {rgb}{0.93,0.93,0.00}
\definecolor {yellow3}             {rgb}{0.80,0.80,0.00}
\definecolor {yellow4}             {rgb}{0.55,0.55,0.00}
\definecolor {gold2}               {rgb}{0.93,0.79,0.00}
\definecolor {gold3}               {rgb}{0.80,0.68,0.00}
\definecolor {gold4}               {rgb}{0.55,0.46,0.00}
\definecolor {goldenrod1}          {rgb}{1.00,0.76,0.15}
\definecolor {goldenrod2}          {rgb}{0.93,0.71,0.13}
\definecolor {goldenrod3}          {rgb}{0.80,0.61,0.11}
\definecolor {goldenrod4}          {rgb}{0.55,0.41,0.08}
\definecolor {darkgoldenrod1}      {rgb}{1.00,0.73,0.06}
\definecolor {darkgoldenrod2}      {rgb}{0.93,0.68,0.05}
\definecolor {darkgoldenrod3}      {rgb}{0.80,0.58,0.05}
\definecolor {darkgoldenrod4}      {rgb}{0.55,0.40,0.03}
\definecolor {rosybrown1}          {rgb}{1.00,0.76,0.76}
\definecolor {rosybrown2}          {rgb}{0.93,0.71,0.71}
\definecolor {rosybrown3}          {rgb}{0.80,0.61,0.61}
\definecolor {rosybrown4}          {rgb}{0.55,0.41,0.41}
\definecolor {indianred1}          {rgb}{1.00,0.42,0.42}
\definecolor {indianred2}          {rgb}{0.93,0.39,0.39}
\definecolor {indianred3}          {rgb}{0.80,0.33,0.33}
\definecolor {indianred4}          {rgb}{0.55,0.23,0.23}
\definecolor {sienna1}             {rgb}{1.00,0.51,0.28}
\definecolor {sienna2}             {rgb}{0.93,0.47,0.26}
\definecolor {sienna3}             {rgb}{0.80,0.41,0.22}
\definecolor {sienna4}             {rgb}{0.55,0.28,0.15}
\definecolor {burlywood1}          {rgb}{1.00,0.83,0.61}
\definecolor {burlywood2}          {rgb}{0.93,0.77,0.57}
\definecolor {burlywood3}          {rgb}{0.80,0.67,0.49}
\definecolor {burlywood4}          {rgb}{0.55,0.45,0.33}
\definecolor {wheat1}              {rgb}{1.00,0.91,0.73}
\definecolor {wheat2}              {rgb}{0.93,0.85,0.68}
\definecolor {wheat3}              {rgb}{0.80,0.73,0.59}
\definecolor {wheat4}              {rgb}{0.55,0.49,0.40}
\definecolor {tan1}                {rgb}{1.00,0.65,0.31}
\definecolor {tan2}                {rgb}{0.93,0.60,0.29}
\definecolor {tan4}                {rgb}{0.55,0.35,0.17}
\definecolor {chocolate1}          {rgb}{1.00,0.50,0.14}
\definecolor {chocolate2}          {rgb}{0.93,0.46,0.13}
\definecolor {chocolate3}          {rgb}{0.80,0.40,0.11}
\definecolor {firebrick1}          {rgb}{1.00,0.19,0.19}
\definecolor {firebrick2}          {rgb}{0.93,0.17,0.17}
\definecolor {firebrick3}          {rgb}{0.80,0.15,0.15}
\definecolor {firebrick4}          {rgb}{0.55,0.10,0.10}
\definecolor {brown1}              {rgb}{1.00,0.25,0.25}
\definecolor {brown2}              {rgb}{0.93,0.23,0.23}
\definecolor {brown3}              {rgb}{0.80,0.20,0.20}
\definecolor {brown4}              {rgb}{0.55,0.14,0.14}
\definecolor {salmon1}             {rgb}{1.00,0.55,0.41}
\definecolor {salmon2}             {rgb}{0.93,0.51,0.38}
\definecolor {salmon3}             {rgb}{0.80,0.44,0.33}
\definecolor {salmon4}             {rgb}{0.55,0.30,0.22}
\definecolor {lightsalmon2}        {rgb}{0.93,0.58,0.45}
\definecolor {lightsalmon3}        {rgb}{0.80,0.51,0.38}
\definecolor {lightsalmon4}        {rgb}{0.55,0.34,0.26}
\definecolor {orange2}             {rgb}{0.93,0.60,0.00}
\definecolor {orange3}             {rgb}{0.80,0.52,0.00}
\definecolor {orange4}             {rgb}{0.55,0.35,0.00}
\definecolor {darkorange1}         {rgb}{1.00,0.50,0.00}
\definecolor {darkorange2}         {rgb}{0.93,0.46,0.00}
\definecolor {darkorange3}         {rgb}{0.80,0.40,0.00}
\definecolor {darkorange4}         {rgb}{0.55,0.27,0.00}
\definecolor {coral1}              {rgb}{1.00,0.45,0.34}
\definecolor {coral2}              {rgb}{0.93,0.42,0.31}
\definecolor {coral3}              {rgb}{0.80,0.36,0.27}
\definecolor {coral4}              {rgb}{0.55,0.24,0.18}
\definecolor {tomato2}             {rgb}{0.93,0.36,0.26}
\definecolor {tomato3}             {rgb}{0.80,0.31,0.22}
\definecolor {tomato4}             {rgb}{0.55,0.21,0.15}
\definecolor {orangered2}          {rgb}{0.93,0.25,0.00}
\definecolor {orangered3}          {rgb}{0.80,0.22,0.00}
\definecolor {orangered4}          {rgb}{0.55,0.15,0.00}
\definecolor {red2}                {rgb}{0.93,0.00,0.00}
\definecolor {red3}                {rgb}{0.80,0.00,0.00}
\definecolor {red4}                {rgb}{0.55,0.00,0.00}
\definecolor {deeppink2}           {rgb}{0.93,0.07,0.54}
\definecolor {deeppink3}           {rgb}{0.80,0.06,0.46}
\definecolor {deeppink4}           {rgb}{0.55,0.04,0.31}
\definecolor {hotpink1}            {rgb}{1.00,0.43,0.71}
\definecolor {hotpink2}            {rgb}{0.93,0.42,0.65}
\definecolor {hotpink3}            {rgb}{0.80,0.38,0.56}
\definecolor {hotpink4}            {rgb}{0.55,0.23,0.38}
\definecolor {pink1}               {rgb}{1.00,0.71,0.77}
\definecolor {pink2}               {rgb}{0.93,0.66,0.72}
\definecolor {pink3}               {rgb}{0.80,0.57,0.62}
\definecolor {pink4}               {rgb}{0.55,0.39,0.42}
\definecolor {lightpink1}          {rgb}{1.00,0.68,0.73}
\definecolor {lightpink2}          {rgb}{0.93,0.64,0.68}
\definecolor {lightpink3}          {rgb}{0.80,0.55,0.58}
\definecolor {lightpink4}          {rgb}{0.55,0.37,0.40}
\definecolor {palevioletred1}      {rgb}{1.00,0.51,0.67}
\definecolor {palevioletred2}      {rgb}{0.93,0.47,0.62}
\definecolor {palevioletred3}      {rgb}{0.80,0.41,0.54}
\definecolor {palevioletred4}      {rgb}{0.55,0.28,0.36}
\definecolor {maroon1}             {rgb}{1.00,0.20,0.70}
\definecolor {maroon2}             {rgb}{0.93,0.19,0.65}
\definecolor {maroon3}             {rgb}{0.80,0.16,0.56}
\definecolor {maroon4}             {rgb}{0.55,0.11,0.38}
\definecolor {violetred1}          {rgb}{1.00,0.24,0.59}
\definecolor {violetred2}          {rgb}{0.93,0.23,0.55}
\definecolor {violetred3}          {rgb}{0.80,0.20,0.47}
\definecolor {violetred4}          {rgb}{0.55,0.13,0.32}
\definecolor {magenta2}            {rgb}{0.93,0.00,0.93}
\definecolor {magenta3}            {rgb}{0.80,0.00,0.80}
\definecolor {magenta4}            {rgb}{0.55,0.00,0.55}
\definecolor {orchid1}             {rgb}{1.00,0.51,0.98}
\definecolor {orchid2}             {rgb}{0.93,0.48,0.91}
\definecolor {orchid3}             {rgb}{0.80,0.41,0.79}
\definecolor {orchid4}             {rgb}{0.55,0.28,0.54}
\definecolor {plum1}               {rgb}{1.00,0.73,1.00}
\definecolor {plum2}               {rgb}{0.93,0.68,0.93}
\definecolor {plum3}               {rgb}{0.80,0.59,0.80}
\definecolor {plum4}               {rgb}{0.55,0.40,0.55}
\definecolor {mediumorchid1}       {rgb}{0.88,0.40,1.00}
\definecolor {mediumorchid2}       {rgb}{0.82,0.37,0.93}
\definecolor {mediumorchid3}       {rgb}{0.71,0.32,0.80}
\definecolor {mediumorchid4}       {rgb}{0.48,0.22,0.55}
\definecolor {darkorchid1}         {rgb}{0.75,0.24,1.00}
\definecolor {darkorchid2}         {rgb}{0.70,0.23,0.93}
\definecolor {darkorchid3}         {rgb}{0.60,0.20,0.80}
\definecolor {darkorchid4}         {rgb}{0.41,0.13,0.55}
\definecolor {purple1}             {rgb}{0.61,0.19,1.00}
\definecolor {purple2}             {rgb}{0.57,0.17,0.93}
\definecolor {purple3}             {rgb}{0.49,0.15,0.80}
\definecolor {purple4}             {rgb}{0.33,0.10,0.55}
\definecolor {mediumpurple1}       {rgb}{0.67,0.51,1.00}
\definecolor {mediumpurple2}       {rgb}{0.62,0.47,0.93}
\definecolor {mediumpurple3}       {rgb}{0.54,0.41,0.80}
\definecolor {mediumpurple4}       {rgb}{0.36,0.28,0.55}
\definecolor {thistle1}            {rgb}{1.00,0.88,1.00}
\definecolor {thistle2}            {rgb}{0.93,0.82,0.93}
\definecolor {thistle3}            {rgb}{0.80,0.71,0.80}
\definecolor {thistle4}            {rgb}{0.55,0.48,0.55}
\definecolor {gray1}               {rgb}{0.01,0.01,0.01}
\definecolor {gray2}               {rgb}{0.02,0.02,0.02}
\definecolor {gray3}               {rgb}{0.03,0.03,0.03}
\definecolor {gray4}               {rgb}{0.04,0.04,0.04}
\definecolor {gray5}               {rgb}{0.05,0.05,0.05}
\definecolor {gray6}               {rgb}{0.06,0.06,0.06}
\definecolor {gray7}               {rgb}{0.07,0.07,0.07}
\definecolor {gray8}               {rgb}{0.08,0.08,0.08}
\definecolor {gray9}               {rgb}{0.09,0.09,0.09}
\definecolor {gray10}              {rgb}{0.10,0.10,0.10}
\definecolor {gray11}              {rgb}{0.11,0.11,0.11}
\definecolor {gray12}              {rgb}{0.12,0.12,0.12}
\definecolor {gray13}              {rgb}{0.13,0.13,0.13}
\definecolor {gray14}              {rgb}{0.14,0.14,0.14}
\definecolor {gray15}              {rgb}{0.15,0.15,0.15}
\definecolor {gray16}              {rgb}{0.16,0.16,0.16}
\definecolor {gray17}              {rgb}{0.17,0.17,0.17}
\definecolor {gray18}              {rgb}{0.18,0.18,0.18}
\definecolor {gray19}              {rgb}{0.19,0.19,0.19}
\definecolor {gray20}              {rgb}{0.20,0.20,0.20}
\definecolor {gray21}              {rgb}{0.21,0.21,0.21}
\definecolor {gray22}              {rgb}{0.22,0.22,0.22}
\definecolor {gray23}              {rgb}{0.23,0.23,0.23}
\definecolor {gray24}              {rgb}{0.24,0.24,0.24}
\definecolor {gray25}              {rgb}{0.25,0.25,0.25}
\definecolor {gray26}              {rgb}{0.26,0.26,0.26}
\definecolor {gray27}              {rgb}{0.27,0.27,0.27}
\definecolor {gray28}              {rgb}{0.28,0.28,0.28}
\definecolor {gray29}              {rgb}{0.29,0.29,0.29}
\definecolor {gray30}              {rgb}{0.30,0.30,0.30}
\definecolor {gray31}              {rgb}{0.31,0.31,0.31}
\definecolor {gray32}              {rgb}{0.32,0.32,0.32}
\definecolor {gray33}              {rgb}{0.33,0.33,0.33}
\definecolor {gray34}              {rgb}{0.34,0.34,0.34}
\definecolor {gray35}              {rgb}{0.35,0.35,0.35}
\definecolor {gray36}              {rgb}{0.36,0.36,0.36}
\definecolor {gray37}              {rgb}{0.37,0.37,0.37}
\definecolor {gray38}              {rgb}{0.38,0.38,0.38}
\definecolor {gray39}              {rgb}{0.39,0.39,0.39}
\definecolor {gray40}              {rgb}{0.40,0.40,0.40}
\definecolor {gray42}              {rgb}{0.42,0.42,0.42}
\definecolor {gray43}              {rgb}{0.43,0.43,0.43}
\definecolor {gray44}              {rgb}{0.44,0.44,0.44}
\definecolor {gray45}              {rgb}{0.45,0.45,0.45}
\definecolor {gray46}              {rgb}{0.46,0.46,0.46}
\definecolor {gray47}              {rgb}{0.47,0.47,0.47}
\definecolor {gray48}              {rgb}{0.48,0.48,0.48}
\definecolor {gray49}              {rgb}{0.49,0.49,0.49}
\definecolor {gray50}              {rgb}{0.50,0.50,0.50}
\definecolor {gray51}              {rgb}{0.51,0.51,0.51}
\definecolor {gray52}              {rgb}{0.52,0.52,0.52}
\definecolor {gray53}              {rgb}{0.53,0.53,0.53}
\definecolor {gray54}              {rgb}{0.54,0.54,0.54}
\definecolor {gray55}              {rgb}{0.55,0.55,0.55}
\definecolor {gray56}              {rgb}{0.56,0.56,0.56}
\definecolor {gray57}              {rgb}{0.57,0.57,0.57}
\definecolor {gray58}              {rgb}{0.58,0.58,0.58}
\definecolor {gray59}              {rgb}{0.59,0.59,0.59}
\definecolor {gray60}              {rgb}{0.60,0.60,0.60}
\definecolor {gray61}              {rgb}{0.61,0.61,0.61}
\definecolor {gray62}              {rgb}{0.62,0.62,0.62}
\definecolor {gray63}              {rgb}{0.63,0.63,0.63}
\definecolor {gray64}              {rgb}{0.64,0.64,0.64}
\definecolor {gray65}              {rgb}{0.65,0.65,0.65}
\definecolor {gray66}              {rgb}{0.66,0.66,0.66}
\definecolor {gray67}              {rgb}{0.67,0.67,0.67}
\definecolor {gray68}              {rgb}{0.68,0.68,0.68}
\definecolor {gray69}              {rgb}{0.69,0.69,0.69}
\definecolor {gray70}              {rgb}{0.70,0.70,0.70}
\definecolor {gray71}              {rgb}{0.71,0.71,0.71}
\definecolor {gray72}              {rgb}{0.72,0.72,0.72}
\definecolor {gray73}              {rgb}{0.73,0.73,0.73}
\definecolor {gray74}              {rgb}{0.74,0.74,0.74}
\definecolor {gray75}              {rgb}{0.75,0.75,0.75}
\definecolor {gray76}              {rgb}{0.76,0.76,0.76}
\definecolor {gray77}              {rgb}{0.77,0.77,0.77}
\definecolor {gray78}              {rgb}{0.78,0.78,0.78}
\definecolor {gray79}              {rgb}{0.79,0.79,0.79}
\definecolor {gray80}              {rgb}{0.80,0.80,0.80}
\definecolor {gray81}              {rgb}{0.81,0.81,0.81}
\definecolor {gray82}              {rgb}{0.82,0.82,0.82}
\definecolor {gray83}              {rgb}{0.83,0.83,0.83}
\definecolor {gray84}              {rgb}{0.84,0.84,0.84}
\definecolor {gray85}              {rgb}{0.85,0.85,0.85}
\definecolor {gray86}              {rgb}{0.86,0.86,0.86}
\definecolor {gray87}              {rgb}{0.87,0.87,0.87}
\definecolor {gray88}              {rgb}{0.88,0.88,0.88}
\definecolor {gray89}              {rgb}{0.89,0.89,0.89}
\definecolor {gray90}              {rgb}{0.90,0.90,0.90}
\definecolor {gray91}              {rgb}{0.91,0.91,0.91}
\definecolor {gray92}              {rgb}{0.92,0.92,0.92}
\definecolor {gray93}              {rgb}{0.93,0.93,0.93}
\definecolor {gray94}              {rgb}{0.94,0.94,0.94}
\definecolor {gray95}              {rgb}{0.95,0.95,0.95}
\definecolor {gray97}              {rgb}{0.97,0.97,0.97}
\definecolor {gray98}              {rgb}{0.98,0.98,0.98}
\definecolor {gray99}              {rgb}{0.99,0.99,0.99}
\definecolor {darkgrey}            {rgb}{0.66,0.66,0.66}

\newcommand{\TODO}[1]{{}}

\newcommand{\ignore}[1]{}

\newcommand{\RSTODO}[1]{{\bf \textcolor{darkgreen}{{\fbox{RS TODO:} #1}}}}
\renewcommand{\RSTODO}[1]{}
\newcommand{\ignoreinshort}[1]{}
 \newcommand{\ignoreinlong}[1]{{#1}}
\def\makenewenumerate#1#2{%
\newcounter{cnt#1}
\newenvironment{#1}%
{\begin{list}{\makebox[0pt][r]{#2}}%
{\setlength{\itemsep}{0pt}% 
 \setlength{\parsep}{.2em}%
 \setlength{\leftmargin}{1.5em}%
 \setlength{\labelwidth}{.4em}%
 \usecounter{cnt#1}}}
{\end{list}}}

\makenewenumerate{myenumerate}{\arabic{cntmyenumerate}.}

\makenewenumerate{renumerate}{\rm(\roman{cntrenumerate})}
\makenewenumerate{renumerateprime}{\rm(\roman{cntrenumerateprime}$'$)}
\makenewenumerate{renumeratesecond}{\rm(\roman{cntrenumeratesecond}$''$)}

\makenewenumerate{aenumerate}{({\it\alph{cntaenumerate}})}
\makenewenumerate{aenumerateprime}{({\it\alph{cntaenumerateprime}$'$})}
\makenewenumerate{aenumeratesecond}{({\it\alph{cntaenumeratesecond}$''$})}

\def\newplaintheorem#1#2{%
\newtheorem{#1plain}{#2}% %% RS: mon mi piace l'indice di sezione
\newenvironment{#1}{\begin{#1plain}\rm }{\end{#1plain}}}

\newplaintheorem{notation}{Notation}
\newplaintheorem{remark}{Remark}
\newplaintheorem{cexample}{Counter-Example}

\newcommand{\noi}{\noindent}

\newcommand{\tuple}[1]{\ensuremath{\langle{#1}\rangle}\xspace}
\newcommand{\set}[1]{\ensuremath{\{{#1}\}}\xspace}
\newcommand{\imp}{\ensuremath{\rightarrow}\xspace}

\renewcommand{\iff}{\ensuremath{\leftrightarrow}\xspace}
\newcommand{\defas}{\ensuremath{\stackrel{\text{\tiny def}}{=}}\xspace}

\newcommand{\pos}{\phantom{\neg}}

\newcommand\cala{\ensuremath{\mathcal{A}}\xspace}

\newcommand\calm{\ensuremath{\mathcal{M}}\xspace}

\newcommand\calt{\ensuremath{\mathcal{T}}\xspace}

\newcommand{\tite}{\ensuremath{{ite}}\xspace}

\newcommand{\mularat}{\ensuremath{\mu_{\larat}}\xspace}

\newcommand\mysout{\bgroup \markoverwith{{-}}\ULon}
\newcommand\nosout{\bgroup \markoverwith{{ }}\ULon}
\definecolor{mygray}{rgb}{0.90,0.90,0.90}
\definecolor{mywhite}{rgb}{1.00,1.00,1.00}

\newcommand{\atoms}[1]{\ensuremath{Atoms(#1)}\xspace}

\newcommand{\T}{\ensuremath{\mathcal{T}}\xspace}

\newcommand{\smttt}[1]{\ensuremath{\text{SMT}(#1)}\xspace}

\newcommand{\euf}{\ensuremath{\mathcal{EUF}}\xspace}

\newcommand{\larat}{\ensuremath{\mathcal{LA}(\mathbb{Q})}\xspace}

\renewcommand{\larat}{\ensuremath{\mathcal{LRA}}\xspace}

\newcommand{\smtlarat}{\smttt{\larat}}

\newcommand{\mathsat}{\textsc{MathSAT}\xspace}

\newcommand{\Bool}{\ensuremath{\mathsf{Bool}}\xspace}

\renewcommand{\RSTODO}[1]{\noindent{\textcolor{blue}{{\fbox{RS TODO:} #1}}}}
\newcommand{\APTODO}[1]{\noindent{\textcolor{green}{{\fbox{AP TODO:} #1}}}}

\renewcommand{\TODO}[1]{\noindent{\textcolor{darkviolet}{{\fbox{TODO:} #1}}}}

\renewcommand{\Bool}{\ensuremath{\mathbb{B}}\xspace}

\newcommand{\myint}[3]{\ensuremath{\int_{#1}^{}\! \mathrm{d}#2 \,#3}\xspace}
\renewcommand{\myint}[3]{\ensuremath{\int_{#1}^{}#3 \ \mathrm{d}#2 }\xspace}
\newcommand{\allx}{\ensuremath{\mathbf{x}}\xspace}
\newcommand{\ally}{\ensuremath{\mathbf{y}}\xspace}

\newcommand{\allA}{\ensuremath{\mathbf{A}}\xspace}
\newcommand{\allB}{\ensuremath{\mathbf{B}}\xspace}

\newcommand{\allAstar}{\ensuremath{\mathbf{A}^*}\xspace}

\newcommand{\allpsi}{\ensuremath{\mathbf{\Psi}}\xspace}

\newcommand{\vi}{\ensuremath{\varphi}}

\newcommand{\vixa}{\ensuremath{\vi(\allx,\allA)}\xspace}

\newcommand{\mua}{\ensuremath{\mu^{\Bool}}\xspace}
\renewcommand{\mua}{\ensuremath{\mu^{\allA}}\xspace}
\renewcommand{\mularat}{\ensuremath{\mu^{\larat}}\xspace}

\newcommand{\muastar}{\ensuremath{\mu^{\allAstar}}\xspace}
\newcommand{\MUASTAR}{{\mathcal M}^{\allAstar}}

\newcommand{\mupsi}{\ensuremath{\mu^{\conditionset}}\xspace}

\newcommand{\vistar}{\ensuremath{\vi^{*}}\xspace}
\newcommand{\vistarstar}{\ensuremath{\vi^{**}}\xspace}
\newcommand{\vistarof}[2]{\ensuremath{\vi^{*}(#1,#2)}\xspace}

\newcommand{\vimuagen}[2]{\ensuremath{#1_{|#2}}\xspace}
\renewcommand{\vimuagen}[2]{\ensuremath{#1_{[#2]}}\xspace}
\newcommand{\vimua}{\vimuagen{\vi}{\mua}}

\newcommand{\vistarmuastar}{\vimuagen{\vistar}{\muastar}}
\newcommand{\vistarstarmuastar}{\vimuagen{\vistarstar}{\muastar}}
\newcommand{\vistarstarmua}{\vimuagen{\vistarstar}{\mua}}

\newcommand{\w}{\ensuremath{w}\xspace}
\newcommand{\wof}[2]{\ensuremath{w(#1,#2)}\xspace}
\newcommand{\wxa}{\wof{\allx}{\allA}}
\newcommand{\wmuagen}[1]{\ensuremath{w_{|#1}}\xspace}
\renewcommand{\wmuagen}[1]{\ensuremath{w_{[#1]}}\xspace}
\newcommand{\wmua}{\wmuagen{\mua}}

\newcommand{\wstar}{\ensuremath{w^{*}}\xspace}
\newcommand{\wstarof}[2]{\ensuremath{w^{*}(#1,#2)}\xspace}

\newcommand{\wstarmuagen}[1]{\ensuremath{w^{*}_{|#1}}\xspace}
\renewcommand{\wstarmuagen}[1]{\ensuremath{w^{*}_{[#1]}}\xspace}
\newcommand{\wstarmuastar}{\wstarmuagen{\muastar}}

\newcommand{\wmuapsi}{\ensuremath{\wmuagen{\mua\mupsi}}}
\renewcommand{\wmuapsi}{\ensuremath{f_{\mua\mupsi}}}

\newcommand{\WMIgen}[4]{\ensuremath{{\sf WMI}(#1,#2|#3,#4)}\xspace}
\newcommand{\WMIviwxa}{\WMIgen{\vi}{\w}{\allx}{\allA}}
\newcommand{\WMINBgen}[3]{\ensuremath{{\sf WMI_{nb}}(#1,#2|#3)}\xspace}

\newcommand{\WMINBviwx}{\WMINBgen{\vi}{\w}{\allx}}

\newcommand{\WMI}{\ensuremath{{\sf WMI}}\xspace}

\newcommand{\predabsg}[2]{\ensuremath{{\sf PredAbs}(#1,#2)}\xspace}
\renewcommand{\predabsg}[2]{\ensuremath{{\sf PredAbs}_{[#1]}(#2)}\xspace}

\newcommand{\TA}[1]{\ensuremath{\calt\hspace{-.1cm}\cala(#1)}\xspace}
\newcommand{\TTA}[1]{\ensuremath{\calt\hspace{-.1cm}\calt\hspace{-.1cm}\cala(#1)}\xspace}

\newcommand{\ti}[1]{\ensuremath{\sf{t}^{(#1)}}\xspace}
\newcommand{\tn}[1]{\ti{n}}

\newcommand{\conditionset}{\ensuremath{\mathbf{\Psi}}}
\newcommand{\supportwff}{\ensuremath{\chi}}

\newcommand{\FI}{\ensuremath{{\sf FI}^{\larat}}\xspace}
\newcommand{\FIUC}{\ensuremath{{\sf FIUC}^{\larat}}\xspace}

\newcommand{\openshortcut}{\llbracket}
\newcommand{\closeshortcut}{\rrbracket}
\newcommand{\shortcut}[1]{\ensuremath{\openshortcut #1\closeshortcut\xspace}}
\newcommand{\inside}[2]{\shortcut{#1\!\in#2{}}}

\renewcommand{\tite}[3]{\shortcut{{\sf If}\ #1\ {\sf Then}\ #2\ {\sf Else}\ #3{}}}

\newcommand{\wencxya}{\ensuremath{\shortcut{y=\w}(\allx\cup\ally,\allA)}}
\newcommand{\wenc}{\ensuremath{\shortcut{y=\w}}}

\newcommand{\eufwenc}{\ensuremath{\shortcut{y=\w}_\euf}}

\newcommand{\wmuastar}{\wmuagen{\mua*}}

\renewcommand{\TODO}[1]{\todo[inline,color=green!40]{{\small{#1}}}}

\newcommand{\und}[1]{\underline{#1}}

\newcommand{\skeleton}[1]{\ensuremath{{\sf sk}(#1)}}
\newcommand{\skw}[1]{\skeleton{\w}}

\newcommand{\newvistar}{\ensuremath{\vi^{**}}\xspace}

\newcommand{\newvistarmu}{\vimuagen{\newvistar}{\mu}}

\newcommand{\myf}[1]{f_{#1}}
\newcommand{\myeuf}[1]{f_{#1}}
\renewcommand{\myeuf}[1]{f_{#1}(\allx)}

\renewcommand{\myf}[1]
{\ifthenelse%
 {\equal{#1}{11}}{x_1^2x_2}%
 {\ifthenelse%
  {\equal{#1}{12}}{x_1^3x_2}
  {\ifthenelse%
   {\equal{#1}{21}}{x_1x_2^2}
   {\ifthenelse%
    {\equal{#1}{22}}{x_1x_2^3}
    {\ifthenelse%
     {\equal{#1}{3}}{2x_1x_2}
     {3x_1x_2}
}}}}}

\newcommand{\wmipa}{\textsc{WMI-PA}\xspace}
\newcommand{\wmipaeuf}{\textsc{SA-WMI-PA}\xspace}

\newcommand{\latteintegrale}{\textsc{LattE Integrale}\xspace}

\newcommand{\MUA}{{\mathcal M}^{\allA}}
\newcommand{\MULARAT}{{\mathcal M}^{\larat}}

\newcommand{\viquery}{\ensuremath{\vi_{{\sf query}}}}
\newcommand{\rngAtom}{\ensuremath{A^{\Bool/\mathbb{R}}}\xspace}

\newcommand{\rngTreeB}[1]{{\sf r}_\vi(#1)}
\newcommand{\rngTreeR}[1]{{\sf r}_w(#1)}

\title{SMT-based Weighted Model Integration with Structure Awareness}

\author[1]{\href{mailto:<giuseppe.spallitta@unitn.it>?Subject=WA-WMIPA}{Giuseppe Spallitta}{}}
\author[1]{Gabriele Masina}
\author[2]{Paolo Morettin}
\author[1]{Andrea Passerini}
\author[1]{Roberto Sebastiani}
\affil[1]{%
    University of Trento
}
\affil[2]{%
    KU Leuven
}
  
\begin{document}
\maketitle

\begin{abstract}
Weighted Model Integration (WMI) is a popular formalism aimed at unifying approaches for probabilistic inference in hybrid domains, involving
logical and algebraic constraints. Despite a considerable amount of
recent work, allowing WMI algorithms to scale with the complexity of
the hybrid problem is still a challenge. In this paper we highlight
some substantial limitations of existing state-of-the-art solutions,
and develop an algorithm that combines SMT-based enumeration, an efficient technique in formal verification, with an effective encoding of the problem structure. 
This allows our algorithm to avoid generating redundant models, resulting in substantial computational savings. An extensive experimental evaluation on both synthetic and
real-world datasets confirms the advantage of the proposed solution
over existing alternatives.
\end{abstract}
%%%%%%%%%%%%%%%%%%%%%%%%%%%%%%%%%%%%%%%%%%%%%%%%%%%%%%%%%%%%%
%%%
%%%%%%%%%%%%%%%%%%%%%%%%%%%%%%%%%%%%%%%%%%%%%%%%%%%%%%%%%%%%%
\section{Introduction}
\label{sec:intro}
\ignoreinshort{There is a growing interest in the artificial
  intelligence community for extending probabilistic reasoning
  approaches to deal with hybrid domains, characterized by both
  continuous and discrete variables and their relationships. Indeed,
  Hybrid domains are extremely common in real-world scenarios, from
  transport modelling~\citep{hensher2007handbook} to probabilistic
  robotics~\citep{Thrun:2005:PR:1121596} and cyber-physical
  systems~\citep{Lee2008}.} Weighted Model
Integration~\citep{belleijcai15} recently emerged as a unifying
formalism for probabilistic inference in hybrid domains\ignoreinlong{,
  characterized by both continuous and discrete variables and their
  relationships}. The paradigm extends Weighted Model Counting
(WMC)~\citep{Chavira2008}, which is the task of computing the weighted
sum of the set of satisfying assignments of a propositional formula,
to deal with SMT formulas~(e.g. \citep{barrettsst09})\ignoreinshort{consisting of
combinations of Boolean variables and connectives with symbols from a
background theory, like linear arithmetic over the rationals (\larat)}.
Whereas WMC can be made extremely efficient by leveraging component
caching techniques~\citep{sang2004cc,bacchus2009solving}, these
strategies are hard to apply for WMI because of the tight coupling
induced by the arithmetic constraints. Indeed, 
\ignoreinshort{existing} 
component
caching approaches for WMI are restricted to fully factorized
densities 
\ignoreinshort{for which eager encoding of the theory lemmas doesn't blow
up}
\ignoreinlong{with few dependencies among continuous variables}
~\citep{BelleAAAI16}. Another direction specifically targets acyclic~\citep{ZengB19,ZengMYVB20} or
loopy~\citep{ZengMYVB20b} pairwise models. \ignoreinshort{Approximations with
guarantees can be computed for problems in disjunctive normal
form~\citep{abboud2020approximability} or, in restricted cases,
conjunctive normal form~\citep{belleuai15}.}

Exact solutions for more general classes of densities and
constraints mainly leverage advancements in SMT
technology or in knowledge
compilation (KC)~\citep{DarwicheM02}. WMI-PA~\citep{morettin-wmi-ijcar17,morettin-wmi-aij19}
relies on SMT-based Predicate Abstraction
(PA)~\citep{allsmt} to \ignoreinshort{both} reduce the number of models
to be generated and integrated over\ignoreinshort{ and efficiently enumerate them},
and was shown to achieve substantial improvements over previous
solutions.
%~\cite{morettin-wmi-aij19}. 
%
However, we show how WMI-PA has the major drawback of ignoring the
structure of the weight function when pruning away redundant
models. This seriously affects its simplification power when dealing
with symmetries in the density.
The use of KC for hybrid probabilistic inference
was pioneered by \cite{SannerA12} and further refined in a series of
later
works~\citep{KolbMSBK18,dos2019exact,kolb2020exploit,feldstein2021lifted}. By
compiling a formula into an algebraic circuit, KC
techniques can exploit the structure of the problem to reduce the size
of the resulting circuit, and are at the core of many state-of-the-art
approaches for WMC~\citep{Chavira2008}.  However, even the most recent
solutions for WMI~\citep{dos2019exact,kolb2020exploit} have serious
troubles in dealing with densely coupled problems, resulting in
exponentially large circuits\ignoreinshort{ that are impossible to store or hard
evaluate}.
%do not yet fully exploit the algebraic structure of the problem, resulting in exponentially large circuits that are impossible to store or hard evaluate.  \APTODO{mmh, direi (anche) qualcosa sulla falsariga di quanto mostriamo come issue dei DD no? RS?}

In this paper we introduce a novel algorithm for WMI that aims to
combine the best of both worlds, by introducing weight-structure
awareness into PA-based WMI.
The main idea is to iteratively build a formula which mimics the
conditional structure of the weight function, so as to drive the
SMT-based enumeration algorithm preventing it from generating
redundant models. An extensive experimental evaluation on synthetic
and real-world datasets shows substantial computational advantages of
the proposed solution over existing alternatives for the most
challenging settings.% with large coupling between variables.  

Our  main contributions can be summarized as follows:

\begin{itemize}
    \item We identify major efficiency issues of existing state-of-the-art WMI approaches, both PA and knowledge-compilation based.
    \item We introduce \wmipaeuf{}, a novel WMI algorithm that combines PA with weight-structure awareness.
    \item We show how \wmipaeuf{} achieves substantial computational improvements over existing solutions in both synthetic and real-world settings.
\end{itemize}

\ignoreinshort{The rest of the paper is organized as follows. In Section 2 we
introduce the background, focusing on the formulation of Weighted
Model Integration. Section 3 sums up the strengths and weaknesses of
WMI approaches based on symbolic integration and on Predicate
Abstraction. In Section 4 we address the gaps mentioned in the
previous section, showing theoretical ideas to make WMI-PA
structure-aware and its implementation into the novel algorithm
\wmipaeuf{}. Section 5 is devoted to an empirical evaluation of
\wmipaeuf{} with respect to the other existing approaches, considering
both synthetic and real-world settings.  At last, our conclusions and
final considerations are presented in Section 6.}

%%%%%%%%%%%%%%%%%%%%%%%%%%%%%%%%%%%%%%%%%%%%%%%%%%%%%%%%%%%%% 
%%%
%%%%%%%%%%%%%%%%%%%%%%%%%%%%%%%%%%%%%%%%%%%%%%%%%%%%%%%%%%%%%
%\section{Related Work}
%\label{sec:related}
%\input{related.tex}

%%%%%%%%%%%%%%%%%%%%%%%%%%%%%%%%%%%%%%%%%%%%%%%%%%%%%%%%%%%%%
%%%
%%%%%%%%%%%%%%%%%%%%%%%%%%%%%%%%%%%%%%%%%%%%%%%%%%%%%%%%%%%%%
\section{Background}
\label{sec:background}
%%%
%%%%%%%%%%%%%%%%%%%%%%%%%%%%%%%%%%%%%%%%%%%%%%%%%%%%%%%%%%%%%
\subsection{SMT and Predicate Abstraction}
\label{sec:background-smt}
Satisfiability Modulo Theories (SMT) (see \cite{barrettsst09}) consists in deciding the
satisfiability of first-order formulas over some given theory. For the
context of this paper, we will refer to quantifier-free SMT formulas
over linear real arithmetic (\larat{}), possibly combined with
uninterpreted function symbols ($\larat{}\cup\euf{}$).
We adopt the notation and definitions in
\cite{morettin-wmi-aij19}.
We use $\Bool \defas \{\top, \perp\}$ to indicate the set of Boolean
value, whereas $\mathbb{R}$ indicates the set of real
values. SMT(\larat{}) formulas combines Boolean variables $A_i \in
\Bool$ and \larat{} atoms in the form $(\sum_i c_ix_i\ \bowtie c)$
(where $c_i$ are rational values, $x_i$ are real variables in
$\mathbb{R}$ and $\bowtie$ is one of the standard algebraic operators
$\{=, \neq, <, >, \leq, \geq\}$)  by using standard Boolean operators
$\{\neg,\wedge,\vee,\rightarrow,\leftrightarrow\}$. In
SMT($\larat\cup\euf$), \larat{} terms can
be interleaved  with uninterpreted function symbols.
\noi
Some shortcuts are provided to simplify the reading. The formula $(x_i
\geq l) \wedge (x_i \leq u)$ is shortened into
$\inside{x_i}{[l,u]}$.
% \RSTODO{riscrivere} The formula $(\vi \rightarrow \psi_1) \wedge (\neg \vi \rightarrow \psi_2)$ is abbreviated into $({\sf If}\ \vi\ {\sf Then}\ \psi_1\ {\sf Else}\ \{\psi_2 \})$.

\noi Given an SMT formula $\vi$, a \textit{total} truth assignment $\mu$ is a function that maps \textit{every} atom in $\vi$ to a truth value in $\Bool$. A \textit{partial} truth assignment maps only a subset of atoms to $\Bool$.

%%%
%%%%%%%%%%%%%%%%%%%%%%%%%%%%%%%%%%%%%%%%%%%%%%%%%%%%%%%%%%%%%
\subsection{Weighted Model Integration (WMI)}
\label{sec:background-wmi}
%
%
%\noindent
% We adopt the notation and definitions in
% \cite{morettin-wmi-aij19}.
Let $\allx \defas \{x_1, ..., x_N\} \in \mathbb{R}^N$ and $\allA
\defas \{A_1, ..., A_M\} \in \Bool^M$ for some integers $N$ and
$M$. $\vi(\allx,\allA)$ denotes an SMT(\larat{}) formula over
variables in \allA{} and \allx{} (subgroup of variables are
admissible), while $\w{}(\allx,\allA)$ denotes a non-negative weight
function s.t. $\mathbb{R}^N\times\Bool^M \longmapsto \mathbb{R}^+$.
Intuitively, $\w{}$ encodes a (possibly unnormalized) density function over $\allA \times \allx$.
Hereafter $\mua$ denotes a truth assignment on \allA,
\mularat{} denotes a truth assignment on the \larat-atoms of $\vi$,
$\vimua(\allx)$ denotes (any formula equivalent to) the formula
obtained from \vi{} by substituting every Boolean value $A_i$ with its
truth value in $\mua$ and propagating the truth values through Boolean
operators, and $\wmua(\allx)\defas\w{}(\allx,\mua)$ is
\w{} restricted to %computed on \allx and
the truth values of \mua. 

Given a theory $\T\in\set{\larat,\larat\cup\euf}$,
the nomenclature $\TTA{\vi} \defas \{\mu_1, ..., \mu_j \}$ defines the
set of \T{}-consistent \textit{total} assignments over both
propositional and \T{} atoms that propositionally satisfy $\vi$;
$\TA{\vi} \defas \{\mu_1, ..., \mu_j \}$ represents one set of \T{}
partial assignments over both propositional and \T{} atoms that
propositionally satisfy $\vi$, s.t. every total assignment in
$\TTA{\vi}$ is a super-assignment of some of the partial ones in
$\TA{\vi}$.
% \\. We remark that, given a formula $\vi$, then $\TTA{\vi}$ is unique, while multiple $\TA{\vi}$ are admissible for the same formula $\vi$ (including $\TTA{\vi}$).
\ignoreinshort{The disjunction of the truth assignments in \TTA{\vi} and that of
\TA{\vi} are \T{}-equivalent to \vi.}
Given by $\vixa$, with $\TTA{\exists \allx. \vi}$ we mean the set of all
total truth assignment \mua{} on \allA 
s.t. $\vimua(\allx)$ is \T{}-satisfiable, and by
$\TA{\exists \allx. \vi}$ a set of partial ones s.t.
every total assignment in
$\TTA{\exists \allx. \vi}$ is a super-assignment of some of the partial ones in
$\TA{\exists \allx. \vi}$.
\ignoreinshort{The disjunction of the truth assignments in \TTA{\exists \allx. \vi} and that of
\TA{\exists \allx. \vi} are \T{}-equivalent to $\exists \allx. \vi$.}

% \RSTODO{Spiegare la relazione tra $\TTA{\exists
%     \allx. \vi}, \TA{\exists \allx. \vi}$ e projected enumeration.
% }

%Given \allx, \allA, \w{}(\allx,\allA), $\vi(\allx,\allA)$,
The {\bf Weighted Model
  Integral} of \w{}(\allx,\allA)  over $\vi(\allx,\allA)$
is defined as follows \citep{morettin-wmi-aij19}:
\begin{eqnarray}
%\textstyle
  \label{eq:wmi12}
%    \nonumber
\ \hspace{-.3cm}  \WMIviwxa
\!\!\!&\defas&\!\!\!\!\!
\sum_{\mua\in \Bool^M}^{} \WMINBgen{\vimua}{\wmua}{\allx},
\\
\label{eq:wmi_decomposition3}
&=& %\hspace{-.5cm}  
    \!\!\!\!\!\!\!\!\!\!\!
    \sum_{\mua\in\TTA{\exists \allx. \vi}} \!\!\!\!\!\!\!\!
\WMINBgen{\vimua}{\wmua}{\allx}
  \\
\hspace{-.3cm}\label{eq:wminb1}
    \WMINBviwx\
 &\defas&
  \myint{\vi(\allx)}{\allx}{\w(\allx)},
\\
%   \\
\label{eq:wminb2}
  &\!\!=\!\!& \!\!\!\!
      \sum_{\mularat\in\TA{\vi}}
%      \WMINBgen{\mularat}{\w}{\allx}.
 \myint{\mularat}{\allx}{\w(\allx)},  
\end{eqnarray}
%\RSTODO{definire (qui o sopra) $\mularat$}
where 
the $\mua$'s are all total truth assignments on \allA,
%``${\sf _{nb}}$'' means ``no-Booleans'', that is, 
\WMINBviwx\ is the integral of $\w{}(\allx)$ over the 
set \set{\allx\ |\ \vi(\allx)\ is\ true} (``${\sf _{nb}}$'' means ``no-Booleans'').

We call a {\em support} of a weight
function \wxa{} any subset of $\mathbb{R}^N\times\Bool^M$ out of which
$\wxa=0$, and we represent it as a \larat-formula
$\supportwff(\allx,\allA)$.
We recall that, consequently,
\begin{eqnarray}
\label{eq:support}
  \WMIgen{\vi\wedge\supportwff}{\w}{\allx}{\allA}=
  \WMIgen{\vi}{\w}{\allx}{\allA}.
\end{eqnarray}
%\newpage
We consider the class of {\em feasibly integrable on \larat (\FI)} functions $\w(\allx)$, which contain no
conditional component, and
for which there exists
some procedure able
to compute \WMINBgen{\mularat}{\w}{\allx} for every set of
\larat literals on \allx. (E.g., polynomials are \FI.)
% Such background procedure, which we use as a blackbox, 
% is the basic building block of our WMI calculations. 
Then we call a weight function \wxa{}, {\em feasibly integrable under \larat
  conditions (\FIUC)} iff it can be described in terms of 
% \begin{itemize}
% \item
  a support \larat-formula $\supportwff(\allx,\allA)$
($\top$ if not provided),
%\item
  a set 
  $\conditionset\defas\set{\psi_i(\allx,\allA)}_{i=1}^K$
  of 
 {\em \larat{} {conditions}}, 
%\end{itemize}
in such a way that,
%for every total truth assignment $\mua$ to \allA and 
for every total %\larat-satisfiable 
truth assignment $\mupsi$ to $\conditionset$, %we have that 
\wmuagen{\mupsi}(\allx) is total and \FI in the domain given by the values of
\tuple{\allx,\allA} which satisfy
$\vimuagen{(\supportwff\wedge\mupsi)}{\mua}$.
\ignore{%%%% useless
We denote such \FI functions by $\wmuapsi(\allx)$, s.t.
 for every $\tuple{\mua,\mupsi}$,
\begin{eqnarray}
  \label{eq:FIUC}
\mbox{if $\mua\wedge\mupsi$ holds, then $\w(\allx)=\wmuagen{\mupsi}(\allx)$}.
\end{eqnarray}
} %%%%%%%%%%%%%%%%%%
%
% \TODO{AP: shouldn't it be $\wmuagen{\mupsi}(\allx)$?}
%% RS: si, corretto
%\RSTODO{aggiungere issue su partial}
\FIUC{} functions are all the
weight functions which %in principle,
 can be described  by means of arbitrary combinations of nested
if-then-else's  on conditions in \allA and \conditionset{}, s.t. each
branch \mupsi{} results into a \FI weight function. 
Each \mupsi{} 
describes a portion of the domain of \w{}, inside which
\wmuagen{\mupsi}(\allx)
is \FI, and we say that \mupsi{}  {\em identifies}
\wmuagen{\mupsi} in \w{}.

% \TODO{aggiungere che \FIUC{} functions sono espresse in termini di
%   operazioni aritmetiche, funzioni aritmetiche e ITEs.}
In what follows 
we assume w.l.o.g. that \FIUC{} functions are described as 
combinations  of constants, variables, standard mathematical operators
$+,-,\cdot,/$ un-conditioned mathematical functions (e.g.,
$exp, sin, ...$), conditional expressions
in the form
$({\sf If}\ \psi_i\ {\sf Then}\ t_{1i}\ {\sf Else}\ {t_{2i}})$ whose conditions
$\psi_i$ are \larat{} formulas and terms $t_{1i},t_{2i}$ are \FIUC{}.
\ignore{\footnote{We assume also that every conditional expression
% in the form
% $({\sf If}\ \psi_i\ {\sf Then}\ t_{1i}\ {\sf Else}\ {t_{2i}})$
occurring in \w{}  is
s.t. $t_{1i},t_{2i}$ are not equivalent. Otherwise, we can safely rewrite $({\sf If}\ \psi_i\ {\sf Then}\ t_{1i}\ {\sf Else}\ {t_{2i}})$ into ${t_{1i}}$.}}
%%
%%%%%%%%%%%%%%%%%%%%%%%%%%%%%%%%%%%%%%%%%%%%%%%%%%%%%%%%%%%%%
%\subsection{WMI via Predicate Abstraction (\wmipa{})}
\subsection{WMI via Predicate Abstraction}
\label{sec:background-wmipa}
%\TODO{BLA BLA}
\wmipa{} is an efficient WMI algorithm presented in \cite{morettin-wmi-ijcar17,morettin-wmi-aij19}
which exploits SMT-based predicate abstraction.
Let \wxa{} be a \FIUC{} function as above.
\wmipa{} is based on the fact that 
\begin{eqnarray} 
\label{eq:wmi3}
%\nonumber
%%\textstyle
\hspace{-.2cm}\WMIviwxa   
\hspace{-.2cm}&=& \hspace{-.9cm}
\sum_{\muastar\in\TTA{\exists \allx. \vistar}} \hspace{-.7cm}
  \WMINBgen{\vistarmuastar}{\wstarmuastar}{\allx}
  \\
  \vistar&\defas&\vi\wedge\supportwff\wedge\bigwedge_{k=1}^{K}(B_k\iff \psi_k)
% \\
  \label{eq:wmi4}
%  \WMINBgen{\vistarmuastar}{\wstarmuastar}{\allx}&\defas&\hspace{-.9cm} \sum_{\mularat\in\TA{\vistarmuastar}}  \hspace{-.5cm}
%  \WMINBgen{\mularat}{\wstarmuastar}{\allx},
\end{eqnarray}
% where each \WMINBgen{\vistarmuastar}{\wstarmuastar}{\allx} %element of the sum 
% can be computed as:
% \begin{eqnarray} 
% \label{eq:wmi4}
% \hspace{-.9cm} \sum_{\mularat\in\TA{\vistarmuastar}}  \hspace{-.5cm}
% \WMINBgen{\mularat}{\wstarmuastar}{\allx},
% \end{eqnarray}
%s.t.
$\allAstar\defas\allA\cup\allB$ s.t. 
$\allB\defas\set{B_1,...,B_K}$ are fresh propositional atoms and 
\wstarof{\allx}{\allA\cup\allB} is the weight function obtained by
substituting in \wxa{} each condition $\psi_k$ with $B_k$.
% , for
% every $k\in [1..K]$.
% , and $\vistar\defas\vi\wedge\supportwff\wedge\bigwedge_{k=1}^{K}(B_k\iff \psi_k)$.

The pseudocode of \wmipa is reported in Algorithm~\ref{alg:wmipa}.
  First, the problem is transformed (if needed) by labeling all conditions
  \allpsi{} occurring in \wxa{} with fresh Boolean variables \allB. After this preprocessing
  stage, the set $\calm^{\allAstar}\defas\TTA{\exists \allx. \vistar}$ is computed by invoking
  SMT-based predicate abstraction
 \cite{allsmt}, namely \TTA{\predabsg{\vistar}{\allAstar}}.
 %   $\TTA{\exists \allx. \vi}, \TA{\exists \allx. \vi}$ can be
 % efficiently computed by SMT-based predicate abstraction
 % \cite{allsmt}. 
 % \PMTODO{What is $\predabsg{}{}$? Do we need it?}
  Then, the algorithm iterates over each
  Boolean assignment \muastar in $\calm^{\allAstar}$.
\vistarmuastar is simplified by the ${\sf Simplify}$
procedure.
Then, if \vistarmuastar %the resulting  formula 
is already a conjunction of literals, 
%no further processing   is needed and 
 the algorithm directly computes its
  contribution to the volume by calling
  $\WMINBgen{\vistarmuastar}{\wstarmuastar}{\allx}$.  Otherwise,
  \TA{\vistarmuastar} is computed as
  \TA{\predabsg{\vistarmuastar}{\atoms{\vistarmuastar}}} to produce partial assignments, and the algorithm iteratively computes
  contributions to the volume for each \mularat.
We refer the reader to  \cite{morettin-wmi-aij19} for more details. 

Notice that in the actual implementation  the potentially-large sets
$\MUASTAR$ and $\MULARAT$ are not generated
explicitly. Rather, their elements are
generated, integrated and then dropped one-by-one, so that to avoid
space blowup.

\begin{algorithm}[t]
\caption{{\sf WMI-PA}($\vi$, $\w$, \allx, \allA)
\label{alg:wmipa}}

\begin{algorithmic}[1]
\STATE $\tuple{\vistar, \wstar, \allAstar} \leftarrow  {\sf LabelConditions}(\vi, \w, \allx, \allA)$
% \STATE return {\sf ComputeVolume}($\vistar$(\allx, $\allA \cup \allB$), $\wstar$)
% \end{algorithmic}
% \end{algorithm}

% \begin{algorithm}[t]
% \caption{{\sf ComputeVolume}($\vi$(\allx, \allA), $\w$)}
% \begin{algorithmic}
\STATE $\MUASTAR \leftarrow \TTA{\predabsg{\vistar}{\allAstar}}$
\STATE $vol \leftarrow 0$
\FOR{$\muastar \in \MUASTAR$}
    \STATE ${\sf Simplify}(\vistarmuastar)$
    \IF{${\sf LiteralConjunction}(\vistarmuastar)$}
    \STATE  $vol \leftarrow vol + \WMINBgen{\vistarmuastar}{\wstarmuastar}{\allx}$
    \ELSE
    \STATE $\MULARAT \leftarrow  \TA{\predabsg{\vistarmuastar}{\atoms{\vistarmuastar}}}$ 
    \FOR{$\mularat \in \MULARAT$}
       \STATE $vol \leftarrow vol + \WMINBgen{\mularat}{\wstarmuastar}{\allx}$
    \ENDFOR
    \ENDIF
\ENDFOR
\STATE return $vol$
\end{algorithmic}
\label{algo:wmi}
\end{algorithm}

%%%%%%%%%%%%%%%%%%%%%%%%%%%%%%%%%%%%%%%%%%%%%%%%%%%%%%%%%%%%% 
%%%
%%%%%%%%%%%%%%%%%%%%%%%%%%%%%%%%%%%%%%%%%%%%%%%%%%%%%%%%%%%%%
\section{Efficiency Issues}
\label{sec:wmipa-issues}
\subsection{Knowledge Compilation}
\label{sec:dd_issues}

% \RSNOTE{Domanda: lo diciamo da qualche parte che gli XADD and (F)XSDD non
% tagliano automaticamente i rami \larat-inconsistenti?}
%\RSTODO{IO la figura la porterei in appendice anziche' eliminarla}
\ignoreinshort{
\begin{figure*}
  \centering
  \includegraphics[width=\textwidth]{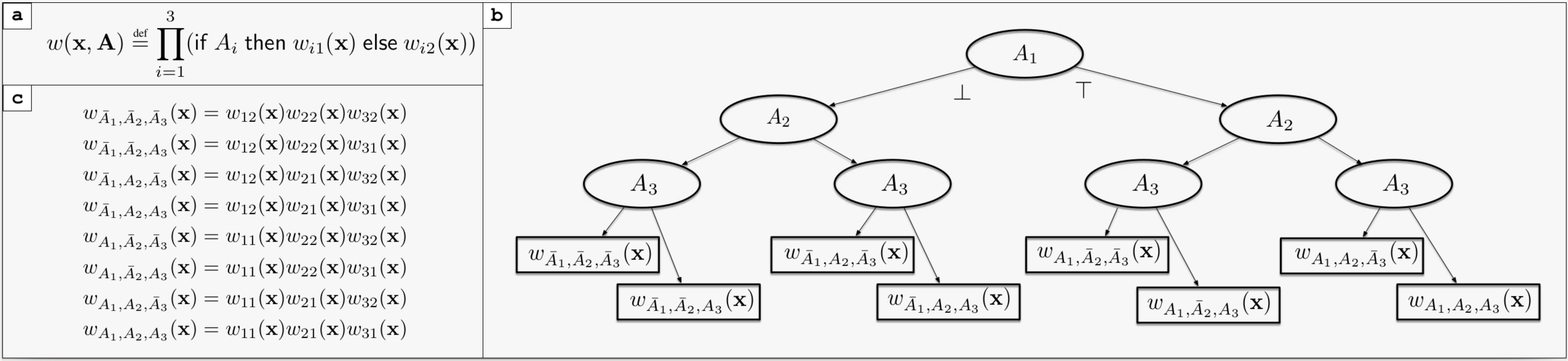}  
  \caption{\label{fig:ex_dd_issues} Example highlighting the
    efficiency issues of the knowledge compilation algorithms for WMI. ({\bf a}) definition
    of a weight function consisting of a product of conditional statements. 
   ({\bf b}) decision diagram generated by knowledge compilation approaches over the weight function. Round nodes
    indicate if-then-else conditions, with true and false cases on the right and left outgoing edges respectively. Squared nodes indicate \FI weight functions. Note how the diagram has a number of branches which is exponential in the number of conditions, and no compression is achieved. ({\bf c}) definition of the weight functions at the leaves of the diagram. In naming weight functions, we use $\bar{A}$ for $\lnot A$ for the sake of compactness.}
\end{figure*}
}

%\TODO{Rseba}
We start our analysis of \WMI{} techniques by noticing a major problem
with existing KC approaches for
WMI~\citep{dos2019exact,kolb2020exploit}, in that they tend easily to
blow up in space even with simple weight functions.  Consider, e.g.,
the case in which
\begin{eqnarray}
  \label{eq:prodite}
  \wxa\defas\prod_{i=1}^N ({\sf if}\ \psi_i\ {\sf then}\ w_{i1}(\allx)\ {\sf else}\ w_{i2}(\allx))
\end{eqnarray}
where the $\psi_i$s are \larat conditions on \set{\allx,\allA} and the
$w_{i1},w_{i2}$ are generic functions on $\allx$.
First, the decision diagrams do not interleave arithmetical and conditional
operators, rather they push all the arithmetic operators below the
conditional ones. Thus with \eqref{eq:prodite} the resulting decision diagrams consist of $2^N$
branches on the $\psi_i$s, each corresponding to a distinct unconditioned weight
function of the form $\prod_{i=1}^Nw_{i{j_i}}(\allx)$ s.t. $j_i\in\set{1,2}$. 
\ignoreinshort{See Figure~\ref{fig:ex_dd_issues} for an example for $N=3$ and
$\psi_i\defas A_i$, $i\in\set{1,2,3}$.}
Second, the decision diagrams are built on the Boolean abstraction of
\wxa, s.t. they do not eliminate a priori the useless
branches consisting in \larat-inconsistent combinations of $\psi_i$s, which can be up to exponentially many. 

With \wmipa{}, instead, 
the representation of \eqref{eq:prodite} does not grow in size,
because \FIUC{} functions allow for interleaving arithmetical and conditional
operators. Also, the SMT-based enumeration algorithm does not generate
\larat-inconsistent assignments on the $\psi_i$s. 
%
%\begin{eqnarray*}
%\wxa\defas\prod_{i=1}^3 ({\sf if}\ A_i\ {\sf then}\ w_{i1}(\allx)\ {\sf else}\ w_{i2}(\allx)) \\
%A_1 \\
%A_2 \\
%A_3 \\
%\top \\
%\bot \\
%w_{\bar{A}_1,\bar{A}_2, \bar{A}_3}(\allx) = w_{12}(\allx)w_{22}(\allx)w_{32}(\allx) \\
%w_{\bar{A}_1,\bar{A}_2, A_3}(\allx) = w_{12}(\allx)w_{22}(\allx)w_{31}(\allx) \\
%w_{\bar{A}_1,A_2, \bar{A}_3}(\allx) = w_{12}(\allx)w_{21}(\allx)w_{32}(\allx) \\
%w_{\bar{A}_1,A_2, A_3}(\allx) = w_{12}(\allx)w_{21}(\allx)w_{31}(\allx) \\
%w_{A_1,\bar{A}_2, \bar{A}_3}(\allx) = w_{11}(\allx)w_{22}(\allx)w_{32}(\allx) \\
%w_{A_1,\bar{A}_2, A_3}(\allx) = w_{11}(\allx)w_{22}(\allx)w_{31}(\allx) \\
%w_{A_1,A_2, \bar{A}_3}(\allx) = w_{11}(\allx)w_{21}(\allx)w_{32}(\allx) \\
%w_{A_1,A_2, A_3}(\allx) = w_{11}(\allx)w_{21}(\allx)w_{31}(\allx) \\
%\end{eqnarray*}
%
We stress the fact that \eqref{eq:prodite} is not an
artificial scenario: rather, e.g., this is the case of the real-world
%road-network 
logistics
problems in \cite{morettin-wmi-aij19}. 

\subsection{WMI-PA}

\begin{figure*}
  \centering
  \includegraphics[width=\textwidth]{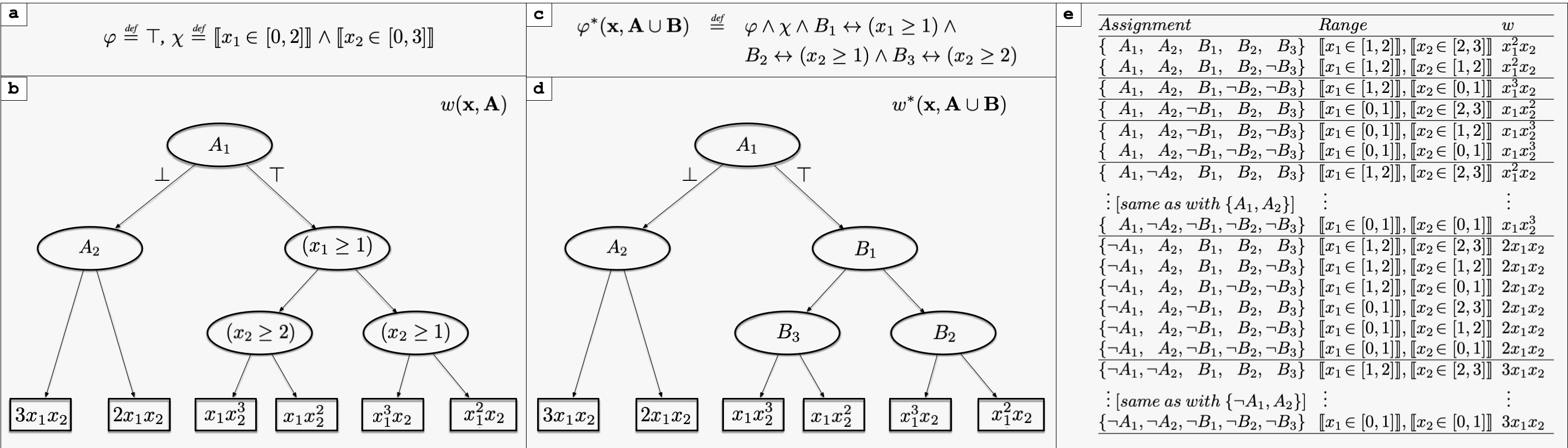}  
  \caption{\label{fig:ex_pa_issues} Example highlighting the
    efficiency issues of the \wmipa{} algorithm. ({\bf a}) definition
    of formula $\vi$ (trivially true) and support $\supportwff$. ({\bf
      b}) definition of the weight function $\wxa$. Round nodes
    indicate if-then-else conditions, with true and false cases on the
    right and left outgoing edges respectively. Squared nodes indicate
    \FI weight functions. ({\bf c}) novel version of the formula
    $\vistarof{\allx}{\allA\cup\allB}$ after the application of the
    {\sf LabelConditions(...)} step of \wmipa{}. ({\bf d}) novel
    version of the weight function $\wstarof{\allx}{\allA\cup\allB}$,
    where all \larat conditions have been replaced with the fresh
    Boolean variables introduced in
    $\vistarof{\allx}{\allA\cup\allB}$. ({\bf e}) List of assignments
    obtained by \wmipa on $\allA\cup\allB$. Notice the amount of
    assignments sharing the same \FI weight function.}
\end{figure*}

% \newpage
We continue our analysis by noticing 
a major deficiency also of the \wmipa{} algorithm, that is,  it fails to leverage the structure of the weight function to prune the set of models to integrate over. We illustrate the issue by means of a simple example (see Figure~\ref{fig:ex_pa_issues}).

\begin{example}
\label{ex:issue1}
Let $\vi\defas\top$,
$\supportwff\defas\inside{x_1}{[0,2]}\wedge\inside{x_2}{[0,3]}$ (Figure~\ref{fig:ex_pa_issues}(a)) and let $\wxa$ be a tree-structured weight function defined as in Figure~\ref{fig:ex_pa_issues}(b).
To compute $\WMIgen{\vi\wedge\supportwff}{\w}{\allx}{\allA}$, six integrals have to be computed:

$\myf{11}$ on $\inside{x_1}{[1,2]}\wedge\inside{x_2}{[1,3]}$ (if $A_1=\top$)\\
$\myf{12}$ on $\inside{x_1}{[1,2]}\wedge\inside{x_2}{[0,1]}$ (if $A_1=\top$)\\
$\myf{21}$ on $\inside{x_1}{[0,1]}\wedge\inside{x_2}{[2,3]}$ (if $A_1=\top$)\\
$\myf{22}$ on $\inside{x_1}{[0,1]}\wedge\inside{x_2}{[0,2]}$ (if $A_1=\top$).\\
$\myf{3}$ on $\inside{x_1}{[0,2]}\wedge\inside{x_2}{[0,3]}$ (if $A_1=\bot,A_2=\top$)\\
$\myf{4}$ on $\inside{x_1}{[0,2]}\wedge\inside{x_2}{[0,3]}$ (if $A_1=\bot,A_2=\bot$)\\

\noi
When \wmipa{} is used (Algorithm~\ref{alg:wmipa}), applying {\sf LabelConditions(...)} we obtain (Figure~\ref{fig:ex_pa_issues}(c)):
%$\vistar\defas\vi\wedge\supportwff\wedge$
%
\begin{eqnarray} 
  \nonumber
  \vistarof{\allx}{\allA\cup\allB}&\defas&\vi\wedge\supportwff\wedge B_1\iff(x_1\ge 1)\wedge\\
  \nonumber
                  &&
                  B_2\iff(x_2\ge 1)\wedge
                  B_3\iff(x_2\ge 2)
\end{eqnarray}
and the weight function $\wstarof{\allx}{\allA\cup\allB}$ shown in Figure~\ref{fig:ex_pa_issues}(d). Then, by applying $\TTA{\predabsg{\vistar}{\allAstar}}$
(row 2)
we obtain 24 total
assignments $\MUASTAR{}$ on $\allA\cup\allB$, as shown in Figure~\ref{fig:ex_pa_issues}(e).\ignoreinshort{(note that assignments containing $\neg B_2,\pos B_3$ are missing as they are theory-inconsistent according to $\vistarof{\allx}{\allA\cup\allB}$)}.
\ignoreinshort{As a result, since each of the 24 \vistarmuastar{}'s is a conjunction of
literals,  \wmipa{}  computes 24 integrals instead of six.} Notice that 
\wmipa{} uselessly splits into 2
parts the integrals on $\myf{11}$ and $\myf{22}$ and into 6 parts the
integral on $\myf{3}$ and on $\myf{4}$. Also, it repeats the very same
integrals for \set{A_1,A_2,...} and  \set{A_1,\neg A_2,...}.
\ignoreinshort{\footnote{The latter fact can be fixed by caching the values of
  the integrals.}}
\hfill $\diamond$
\end{example}

We highlight two facts.
First,
\wmipa{} enumerates {\em total} truth assignments on the Boolean
atoms $\allA\cup\allB$ in  \TTA{\exists \allx. \vistar} \eqref{eq:wmi3} (row 2 in
Algorithm~\ref{algo:wmi}),
assigning also unnecessary values.
Second, \wmipa{} labels \larat conditions in \w{} by means of fresh
Boolean atoms \allB{} (row 1 in Algorithm~\ref{algo:wmi}).
This forces the enumerator to assign all their values in every
assignment, even when not necessary. 

The key issue about \wmipa{} is that the enumeration
of \TTA{\exists \allx. \vistar} in %\eqref{eq:wmi_decomposition3} and
\eqref{eq:wmi3}
and of \TA{\vistarmuastar} in \eqref{eq:wminb2}
(rows 2 and  11
in Algorithm~\ref{algo:wmi})
{\em is not aware of the conditional structure of the weight function
  \w{}},
in particular, it is not aware of the fact that often
{\em partial}
assignments to the set of conditions in \wstar{} (both Boolean and \larat) are sufficient to identify the
value of a \FIUC{} function
(e.g \set{ A_1,B_1,B_2} suffices to identify $\myf{11}$, or \set{\neg A_1,A_2} suffices to identify $\myf{3}$),
so that it is forced
to  enumerate all total assignments extending them
(e.g. \set{A_1,A_2,B_1,B_2,B_3} and \set{A_1,A_2,B_1,B_2,\neg B_3}).

Thus, to cope with this issue, we need to modify \wmipa{} to
make it aware of the conditional structure of \w{}. 
%\TODO{We need to make it structure aware}
%\TODO{Add esample from blackboard}
% \TODO{Explain: \\need TTA for Boolean atoms, \\labeling decisions in cost}
% \TODO{need to make it structure aware}

%%%%%%%%%%%%%%%%%%%%%%%%%%%%%%%%%%%%%%%%%%%%%%%%%%%%%%%%%%%%%
%%%
%%%%%%%%%%%%%%%%%%%%%%%%%%%%%%%%%%%%%%%%%%%%%%%%%%%%%%%%%%%%%
\section{Making WMI-PA Weight-Structure Aware}
\label{sec:wmipa-improved}
%\subsection{Theoretical Ideas}

\ignoreinshort{
Let \wxa{} be \FIUC on the set of conditions $\allpsi$. We notice a couple of facts.
\TODO{Merge next two facts into previous section?}
First,
in \eqref{eq:wmi_decomposition3} and in \eqref{eq:wmi3} the $\mua$s
are {\em total} (that is, 
$\TTA{\exists \allx. \vi}$ is used instead of $\TA{\exists \allx. \vi}$)
because the enumerator does not know $\wxa{}$, which in the general case  needs {\em total} assignments on \allA{} 
even when a {\em partial} one suffices to
make $\vixa$ true. 
Second,
the integral in \eqref{eq:wminb2} is
not straightforward to compute because the computation should be
partitioned by the if-then-else structure of \w{}, which requires a
further form of search on the conditions $\conditionset$ of \w{}.
This forces \wmipa{} to introduce the fresh variables \allB and to
conjoin the constraints $(B_i\iff\psi_i)$ to the formulas, which in turn
forces the assignments to be total on \allB.
}

The key idea to prevent total enumeration works as follows.
We do {\em not} rename with \allB{} the conditions \allpsi in \w{} and,
rather than enumerating total truth assignments for $\exists
\allx.\vistar$
as in \eqref{eq:wmi3}--\eqref{eq:wmi4},
we enumerate {\em partial} assignments for
$\exists \allx. \newvistar$ where $\newvistar\defas\vi\wedge\supportwff\wedge\skw{}$
%${\exists\allx. (\vi\wedge\supportwff\wedge\skw{})}$,
and $\skw{}$ --which we call the {\em conditional skeleton} of $\w$-- is a \larat
formula s.t.:\\
$(a)$ its atoms are all and only the conditions in $\allpsi$,\\
$(b)$ is \larat-valid, so that $\vi\wedge\supportwff$ is equivalent to
$\vi\wedge\supportwff\wedge\skw{}$,\\
$(c)$ any {\em partial} truth value assignment $\mu$ to the conditions
$\allpsi$ which makes \skw{} true is such that \wmuagen{\mu} is \FI{}.%
\footnote{E.g., the partial assignment
  $\mu\defas\set{A_1,(x_1\ge 1),(x_2\ge 1)}$ in
  Example~\ref{ex:issue1} is such that
  $\wmuagen{\mu}=\myf{11}$,
  which is \FI{}.}  \\
%
% (d) non e' una proprieta' delle dafinizione di \skw{}
%$(d)$ does not blow up in size wrt. the size of $\w$.\\
\noindent
Thus, we have that \eqref{eq:wmi_decomposition3} can be rewritten as:
\begin{eqnarray} 
\label{eq:newwmi1}
%\nonumber
%%\textstyle
\hspace{-.5cm}\WMIviwxa   
\hspace{-.3cm}&=& \hspace{-1cm}
\sum_{\mu\in\TA{\exists \allx. \newvistar}} \hspace{-.6cm} 2^{|\allA\setminus\mu|}\cdot
  \WMINBgen{\newvistarmu}{\wmuagen{\mu}}{\allx}
  \\
  \newvistar&\defas&\vi\wedge\supportwff\wedge\skw{}
% \\
  \label{eq:newwmi2}
\end{eqnarray}
\noindent
where $|\allA\setminus\mu|$ is the number of Boolean atoms in \allA
that are not assigned by $\mu$.
Condition $(c)$ guarantees that
$\WMINBgen{\newvistarmu}{\wmuagen{\mu}}{\allx}$ in \eqref{eq:newwmi1}
can be directly computed, without further partitioning. 
The $2^{|\allA\setminus\mu|}$ factor in \eqref{eq:newwmi1} resembles
the fact that, if some Boolean atom $A_i\in\allA$ is not assigned in 
$\mu$,
then $\WMINBgen{\newvistarmu}{\wmuagen{\mu}}{\allx}$ should be counted
twice because $\mu$ represents two assignments $\mu\cup\set{A_i}$ and
$\mu\cup\set{\neg A_i}$ which would produce two identical integrals.

Notice that logic-wise \skw{} is non-informative because it is a valid formula.
Nevertheless, the role of \skw{} is to mimic the structure of $\w$ so that to ``make  the
enumerator aware of the presence of the conditions $\allpsi{}$'',
forcing every assignment $\mu$ 
to assign truth values also to these conditions which are necessary to make
$\wmuagen{\mu}$ \FI{} and hence make
$\WMINBgen{\newvistarmu}{\wmuagen{\mu}}{\allx}$ directly computable,
without further partitioning.
%\RSTODO{The role of \skw{} is twofold...}

%Notice that point $(d)$ is potentially problematic.
An important issue  is to avoid \skw{} blow up in size.
E.g., one could use
as \skw{} a formula encoding
the conditional structure of an XADDs or (F)XSDDs, but this may
cause a blow up in size, as discussed in Section~\ref{sec:dd_issues}.

In order to prevent such problems, we do not generate \skw{}
explicitly. Rather, we build it as a disjunction of
partial assignments over \allpsi{} which we enumerate progressively.
To this extent, we define $\skw{}\defas\exists\ally.\wenc{}$ where 
\wenc{} is a formula on \allA,\allx,\ally
s.t. $\ally\defas\set{y,y_1,...,y_k}$ is a set of
fresh variables.
Thus, 
$\TA{\exists \allx.(\vi\wedge\supportwff\wedge\exists\ally.\wenc{})}$
can be computed as 
$\TA{\exists \allx\ally.(\vi\wedge\supportwff\wedge\wenc{})}$
because the \ally{}'s do not occur in $\vi\wedge\supportwff$,
with no need to generate \skw{} explicitly.
The enumeration of \TA{\exists
  \allx\ally.(\vi\wedge\supportwff\wedge\wenc{})} is performed by
the very same SMT-based procedure used in \cite{morettin-wmi-aij19}.

\wenc{} is obtained by taking $(y=\w)$, s.t. $y$ is fresh, and
recursively substituting bottom-up every
conditional term $({\sf If}\ \psi_i\ {\sf Then}\ t_{i1}\ {\sf Else}\
{t_{i2}})$ 
in it with a fresh variable $y_i\in\ally$, adding the
definition of 
$(y_i=({\sf If}\ \psi_i\ {\sf Then}\ t_{i1}\ {\sf Else}\
{t_{i2}}))$ as
\begin{eqnarray}
  \label{eq:ite}
 (\neg \psi_i\vee y_i=t_{i1})\wedge (\psi_i\vee
y_i=t_{i2}).
\ignore{\footnote{\RSTODO{Spostare dopo.}\\For issues of efficiency, in the actual encoding
  we also conjoin the mutual exclusion condition
  $(\neg(y_i=t_{i1})\vee \neg (y_i=t_{i2}))$ under the assumption that
  $t_{i1},t_{i2}$ are not equivalent. (Otherwise, we can safely
  rewrite
  $({\sf If}\ \psi_i\ {\sf Then}\ t_{i1}\ {\sf Else}\ {t_{i2}})$ into
  ${t_{i1}}$.)  We omit this in the rest of the paper for simplicity
  of notation.}}
\end{eqnarray}
% $(\neg \psi_i\vee y_i=t_{i1})\wedge (\psi_i\vee
% y_i=t_{i2})$.
%
This labeling\&rewriting process, which is inspired to labeling CNF-ization
\citep{tseitin68}, guarantees that the size of \wenc{} is linear wrt. that of \w{}.
E.g., if \eqref{eq:prodite} holds, then
$\wenc$ is $(y=\prod_{i=1}^N
y_i)\wedge\bigwedge_{i=1}^N((\neg\psi_i\vee
y_i=w_{i1}(\allx))\wedge(\psi_i\vee y_i=w_{i2}(\allx)))$. 

%%%%%%%%%%%%%%%%%%%%%%%%%%%%%%%%%%%%%%%%%%%%%%%%%%%%%%%%%

%
% AP: material for Figure~\ref{fig:sa-wmi-pa-labeling} 
%
% \begin{align*}
% (\neg A_1 \vee \neg (x_1\ge 1)\vee \neg (x_2\ge 1) \vee (y_1=\myf{11})) \\
% \wedge 
% (\neg A_1 \vee \neg (x_1\ge 1)\vee \pos (x_2\ge 1) \vee (y_1=\myf{12})) \\
% \end{align*}

% \begin{align*}
% (\neg A_1 \vee \pos (x_1\ge 1)\vee \neg (x_2\ge 2) \vee (y_2=\myf{21})) \\
% \wedge
% (\neg A_1 \vee \pos (x_1\ge 1)\vee \pos (x_2\ge 2) \vee (y_2=\myf{22})) \\
% \end{align*}

% \begin{align*}
% (\neg A_1 \vee \neg (x_1\ge 1) \vee (y_3=y_1)) \\
% \wedge
% (\neg A_1 \vee \pos (x_1\ge 1) \vee (y_3=y_2)) \\
% \end{align*}

% \begin{align*}
% (\pos A_1 \vee \neg A_2 \vee (y_4=\myf{3})) \\
% \wedge
% (\pos A_1 \vee \pos A_2 \vee (y_4=\myf{4})) \\
% \end{align*}

% \begin{align*}
% (\neg A_1 \vee (y_5=y_3)) \\
% \wedge
% (\pos A_1 \vee (y_5=y_4)) \\
% \end{align*}

% $$
% (y=y_5)
% $$

% $$
% y_1 \quad y_2 \quad y_3 \quad y_4 \quad y_5
% $$

%\input{playground}

One problem with the above definition of \wenc{} is that it is not a
\larat-formula, because $\w$ may include multiplications or even
transcendental functions out of the conditions \allpsi\footnote{The
  conditions in \allpsi contain only linear terms by definition.}, which makes SMT reasoning over it
dramatically hard or even undecidable. 
We notice, however, that when computing 
$\TA{\exists \allx\ally.(\vi\wedge\supportwff\wedge\wenc{})}$
the arithmetical functions  (including operators $+,-,*,/$)
occurring in \w{} out of the conditions \allpsi{} have no
role, since the only fact that we need to guarantee for the validity
of \skw{} is that they are indeed functions, so that $\exists
y.(y=f(...))$ is always valid. 
\footnote{This propagates down on the recursive structure of
  \wenc{} because,  if $y$ does not occur in $\psi$,
$\exists y.(y=({\sf If}\ \psi\ {\sf Then}\ t_{1}\ {\sf Else}\
{t_{2}}))$ is equivalent to
$(({\sf If}\ \psi\ {\sf Then}\ \exists y.(y=t_{1})\ {\sf Else}\
{\exists y.(y=t_{2})}))$, and 
$\vi$ is equivalent to $\exists y.(\vi[t|y]\wedge(y=t))$.
}
(In substance, during the enumeration we are interested only in the
truth values  of the conditions $\allpsi$ in $\mu$ which make
$\wmuagen{\mu}$ \FI{},
regardless the actual values of $\wmuagen{\mu}$).
Therefore we can safely substitute condition-less arithmetical
functions (including operators $+,-,\cdot,/$) with some fresh uninterpreted function
symbols, obtaining a $\larat\cup\euf$-formula \eufwenc{}, 
 which
is relatively easy to solve by standard SMT solvers
\citep{barrettsst09}.
It is easy to see that a partial assignment $\mu$ evaluating \wenc{} to
true is \larat-satisfiable iff its corresponding assignment $\mu_\euf$
is $\larat\cup\euf$-satisfiable.\footnote{This boils down to the fact
  that $y$ occurs only in the top equation and as such it is free to assume
  arbitrary values, and that all arithmetic functions are total in the
  domain so that, for every possible values of $\allx$ a value for 
$\ally$ always exists iff there exists in the \euf{} version.}
Therefore, we can modify the
enumeration procedure into
$\TA{\exists \allx\ally.(\vi\wedge\supportwff\wedge\eufwenc{})}$. 

Finally, we enforce the fact  that the two branches of an if-then-else
are alternative by adding to \eqref{eq:ite} a mutual-exclusion
constraint $\neg(y_i=t_{i1})\vee\neg(y_i=t_{i2})$, so that
the choice of the function  is univocally associated to the list of
decisions on the $\psi_i$s.
The   procedure producing \eufwenc{} is
described in detail in Appendix, Algorithm 1. 

% \begin{figure*}
%   \centering
% %  \includegraphics[width=\textwidth]{figures/sa-wmi-pa-labeling_compressed.pdf}
%   \includegraphics[width=\textwidth]{figures/sa-wmi-pa-labeling_screenshot.pdf}
%   \caption{\label{fig:sa-wmi-pa-labeling} Example of bottom-up procedure for computing the relabeling function \eufwenc{}. ({\bf a}) Replacement of 
% \FI{} weight functions (the leaves of the tree, highlighted in red) with \euf function symbols (we dropped the dependency on $x$ for compactness). ({\bf b-g}) Sequence of relabeling steps. At each step, a conditional term is replaced by a fresh \larat{} variable $y_i$. The encoding of the variable in shown in the upper part, while the lower part shows the weight function with the branch of the conditional term replaced with $y_i$ (highlighted in red). The last step consists in renaming the top variable as $y$, so that $y=\wxa{}$. The relabeling function \eufwenc{} is simply the conjunction of the encodings in the different steps.}
% \end{figure*}

\begin{figure*}[!t]
  \centering
 \includegraphics[width=\textwidth]{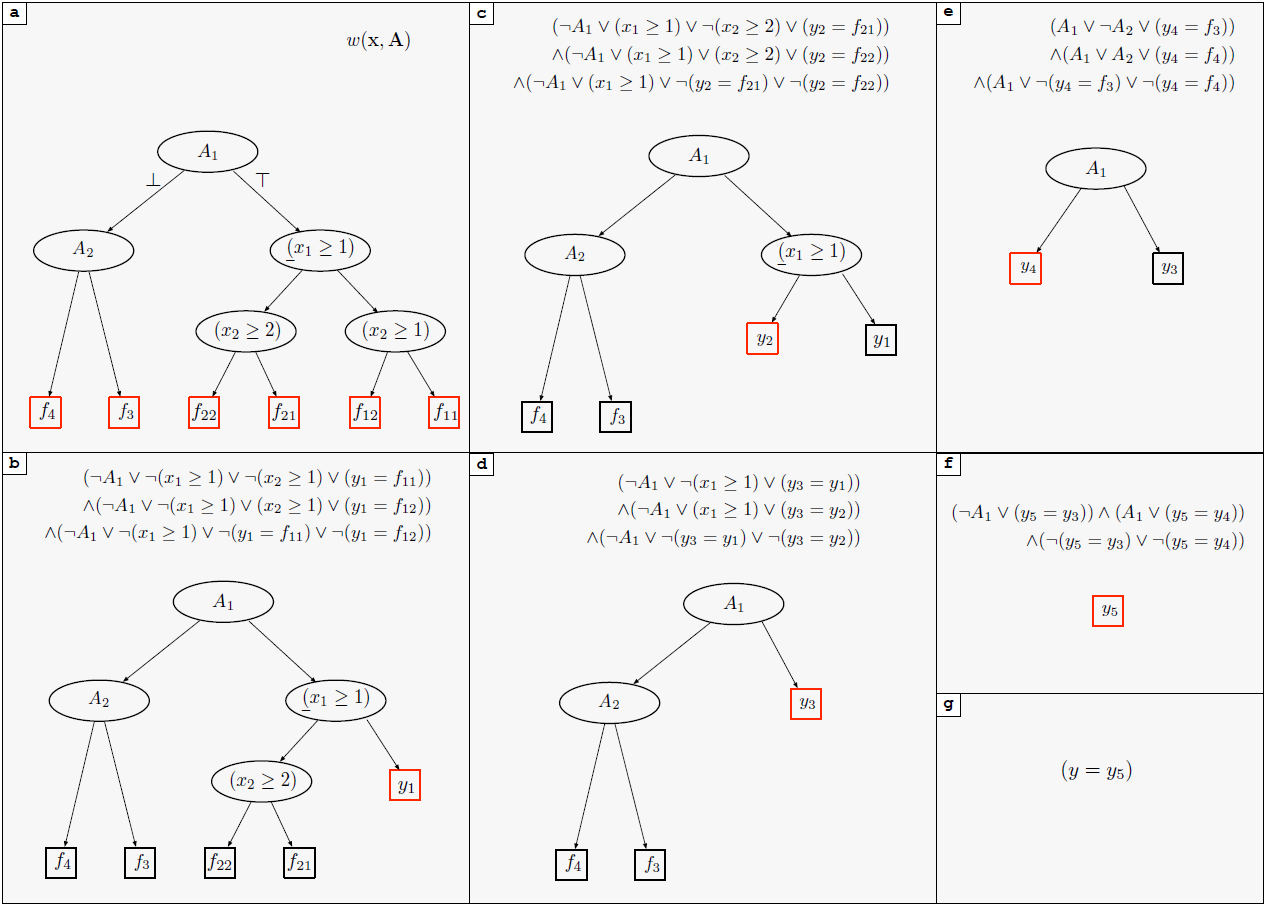}
  \caption{\label{fig:sa-wmi-pa-labeling} Example of bottom-up procedure for computing the relabeling function \eufwenc{}. ({\bf a}) Replacement of 
\FI{} weight functions (the leaves of the tree, highlighted in red) with \euf function symbols (we dropped the dependency on $x$ for compactness). ({\bf b-g}) Sequence of relabeling steps. At each step, a conditional term is replaced by a fresh \larat{} variable $y_i$. The encoding of the variable in shown in the upper part, while the lower part shows the weight function with the branch of the conditional term replaced with $y_i$ (highlighted in red). The last step consists in renaming the top variable as $y$, so that $y=\wxa{}$. The relabeling function \eufwenc{} is simply the conjunction of the encodings in the different steps.}
\end{figure*}

\begin{figure*}[!t]
\centering
 \includegraphics[width=\textwidth]{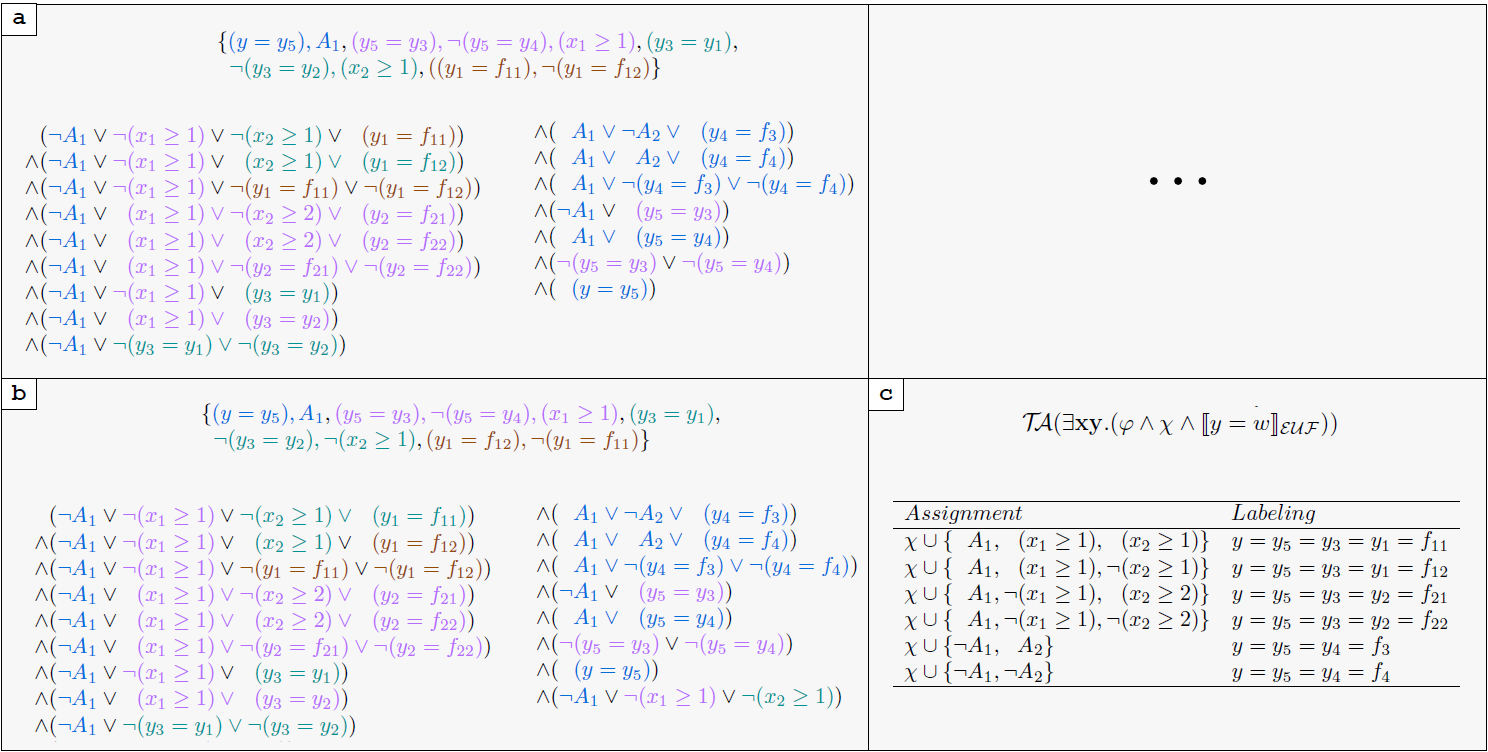}
  \caption{\label{fig:sa-wmi-pa-enumeration} Example of
    structure-aware enumeration performed by \wmipaeuf{} on the
    problem in Example~\ref{ex:issue1}. ({\bf a}) Generation of the
    first assignment. The assignment is on top, while the bottom part
    shows the \eufwenc{} formula. Colors indicate the progression of
    the generation, in terms of atoms added (top) and parts of the
    formula to be removed as a consequence (bottom). For the sake of
    simplicity all atoms until the next atom in
    $\Psi\defas\set{A_1,A_2,(x_1\ge 1),(x_2\ge 1),(x_2\ge 2)}$ (if
    any) are given the same color.  ({\bf b}) Generation of the second
    assignment. Note how the \eufwenc{} formula is enriched with the
    blocking clause $\neg A_1\vee\neg (x_1\ge 1)\vee \neg (x_2\ge 1)$
    preventing the first assignment to be generated again. ({\bf c})
    Final result of the enumeration (which contains six assignments in
    total). The partial assignments are obtained by restricting the
    generated assignments on the conditions in $\Psi$ (and combining
    them with the atoms of $\supportwff$, which here are assigned
    deterministically.).
    For each assignment, the
    corresponding chain of equivalences of the $\ally$s with the
    identified leaf \FI{} function is displayed.}
    % of the $y$ variables is also
    % shown. \RSNOTE{non capisco il senso di quest'ulltima frase. ??}\AP{serve a spiegare la presenza della colonna Labeling..}} 
\end{figure*}

% \begin{example}
% \label{ex:encoding}
% Consider the problem in Example~\ref{ex:issue1}.
% Figure~\ref{fig:sa-wmi-pa-labeling} shows the relabeling process
% applied to the weight function $\w$.  The resulting \eufwenc{} formula
% is:

\begin{example}
\label{ex:encoding}
\label{ex:enumeration}
Consider the problem in Example~\ref{ex:issue1}.
Figure~\ref{fig:sa-wmi-pa-labeling} shows the relabeling process
applied to the weight function $\w$.  The resulting \eufwenc{} formula
is:

{%% come clausole
  % \TODO{RS: togliere mutex?}
  % \TODO{AP: se capisco cosa intendi direi di si', e metterei una
  % footnote spiegando lo facciamo per readability, e che la versione
  % con mutex la lasciamo ai supplementary material}
  %% RS: "Togliere Mutex" era una nota per me stesso :-)
  \hspace{-.2cm}$\begin{array}{ll}
   \phantom{\wedge} %\vi\wedge\supportwff\\
   \wencxya{}\defas\\
   \phantom{\wedge}
   (\neg A_1 \vee \neg (x_1\ge 1)\vee \neg (x_2\ge 1) &\hspace{-.9cm}\vee \pos(y_1=\myeuf{11})) \\
\wedge(\neg A_1 \vee \neg (x_1\ge 1)\vee \pos (x_2\ge 1) &\hspace{-.9cm}\vee \pos(y_1=\myeuf{12})) \\
\wedge(\neg A_1 \vee \neg (x_1\ge 1)\vee \neg (y_1=\myeuf{11})&\hspace{-.9cm}\vee\neg (y_1=\myeuf{12})) \\
\wedge(\neg A_1 \vee \pos (x_1\ge 1)\vee \neg (x_2\ge 2) &\hspace{-.9cm}\vee \pos(y_2=\myeuf{21})) \\
\wedge(\neg A_1 \vee \pos (x_1\ge 1)\vee \pos (x_2\ge 2) &\hspace{-.9cm}\vee \pos(y_2=\myeuf{22})) \\
\wedge(\neg A_1 \vee \pos (x_1\ge 1)\vee \neg (y_2=\myeuf{21})&\hspace{-.9cm}\vee\neg (y_2=\myeuf{22})) \\   
\wedge(\neg A_1 \vee \neg (x_1\ge 1) &\hspace{-.9cm}\vee \pos(y_3=y_1)) \\
\wedge(\neg A_1 \vee \pos (x_1\ge 1) &\hspace{-.9cm}\vee \pos(y_3=y_2)) \\
\wedge(\neg A_1 \vee \neg (y_3=y_1)\vee\neg (y_3=y_2)) &\\   
\wedge(\pos A_1 \vee \neg A_2 &\hspace{-.9cm}\vee \pos(y_4=\myeuf{3})) \\
\wedge(\pos A_1 \vee \pos A_2 &\hspace{-.9cm}\vee \pos(y_4=\myeuf{4})) \\
\wedge(\pos A_1 \vee \neg (y_4=\myeuf{3})\vee\neg (y_4=\myeuf{4})) &\\   
\wedge(\neg A_1 &\hspace{-.9cm}\vee \pos(y_5=y_3)) \\
\wedge(\pos A_1 &\hspace{-.9cm}\vee \pos(y_5=y_4)) \\
\wedge(\neg (y_5=y_3)\vee\neg (y_5=y_4)) &\\   
\wedge(\pos (y=y_5)) &\\   
\end{array}$
}

Figure~\ref{fig:sa-wmi-pa-enumeration}
% illustrates the enumeration process of \wmipaeuf.
% RS: non abbiamo ancora parlato di \wmipaeuf, e inoltre il processo
% non e' univoco, dipende dalle scelte fatte dall'SMT solver.
illustrates a possible enumeration process.
%
%
% \begin{example}
% \label{ex:enumeration}
% Consider the case of
% Example~\ref{ex:encoding}. Figure~\ref{fig:sa-wmi-pa-enumeration}
% % illustrates the enumeration process of \wmipaeuf.
% % RS: non abbiamo ancora parlato di \wmipaeuf, e inoltre il processo
% % non e' univoco, dipende dalle scelte fatte dall'SMT solver.
% illustrates a possible enumeration process.
The algorithm
enumerates partial assignments satisfying
$\vi\wedge\supportwff\wedge\eufwenc{}$, restricted on the conditions
$\Psi\defas\set{A_1,A_2,(x_1\ge 1),(x_2\ge 1),(x_2\ge 2)}$, which is
equivalent to enumerate
$\TA{\exists \allx \ally. (\vi\wedge\supportwff\wedge\eufwenc{})}$.
% Suppose the enumeration algorithm is instructed to decide according
% the following order:
% \set{A_1,A_2,(x_1\ge 1),(x_2\ge 1),(x_2\ge  2)...}.
Assuming 
the enumeration procedure picks nondeterministic choices  
following the order of the above set~\footnote{Like in Algorithm~\ref{alg:wmipaeuf},
 we pick Boolean conditions first.}
and
%$A_1,A_2,(x_1\ge 1),(x_2\ge 1),(x_2\ge  2),...$
assigning positive
values first, then in the first branch the following satisfying partial
assignment is generated, in order: 
\footnote{Here nondeterministic choices are \und{underlined}. The
  atoms in $\supportwff$ are assigned deterministically.}
%\hspace{-.4cm}
$
\begin{array}{l}\supportwff\cup
 \{(y=y_5),
  \und{A_1},(y_5=y_3),\neg(y_5=y_4),
%  \und{A_2}, %%% E' un casino da spiegare
  \und{(x_1\ge 1)},\\(y_3=y_1),\neg(y_3=y_2),
  \und{(x_2\ge 1)},(y_1=\myeuf{11}),\\\neg(y_1=\myeuf{12})\}.
\end{array}
$

(Notice that, following the chains of true equalities, we have $y=y_5=y_3=y_1=\myeuf{11}$.)
Then the SMT solver extracts from it  the subset \set{A_1,(x_1\ge 1), (x_2\ge 1)}
restricted on the conditions in $\Psi$.
\ignore{%%%% TROPPO UN CASINO DA SPIEGARE
  that had an actual role in satisfying the
formula (i.e., $A_2$ is not considered)\footnote{This is due to
  {\em conflict analysis}, a process SMT solvers inherit from SAT. Notice
  that if also the atom
  $(x_2\ge 2)$ were assigned at the end of the assignment, it would have
  ignored like $A_2$ because it would have  no role in satisfying the formula.\APTODO{@RS: replace with ``Notice
  that if the atom
  $(x_2\ge 2)$ were also included at the end of the assignment, it would have
  been ignored (like $A_2$) for having no role in satisfying the formula.''} 
 We refer the reader to
 \cite{allsmt} for more details.}
} %%%%%%%%%%%%%%%%%%%%%%%%%%%%%%%%%%%
Then  the blocking clause $\neg
A_1\vee\neg (x_1\ge 1)\vee \neg (x_2\ge 1)$ is added to the formula, which prevents
to enumerate the same subset again.
This forces the algorithm to backtrack and generate
%\APTODO{@RS: anzi, dovrebbe essere $\und{(x_1\ge 1)}$, no?}. RS: SI
$
\begin{array}{l}\supportwff\cup
  \{(y=y_5),
  \und{A_1},(y_5=y_3),\neg(y_5=y_4),
%  \und{A_2},
  (x_1\ge 1),\\(y_3=y_1),\neg(y_3=y_2),
  {\neg(x_2\ge 1)},(y_1=\myeuf{12}),\\\neg(y_1=\myeuf{11})\}.
\end{array}
$
producing the assignment: \set{A_1,(x_1\ge 1), \neg(x_2\ge 1)}.~%
\footnote{We refer the reader to \cite{allsmt} for more details on the
  SMT-based enumeration algorithm.}

Overall, the algorithm
enumerates the following ordered collection of partial assignments
restricted to $Atoms(\vi\wedge\supportwff)\cup\allpsi$:\\
$
\begin{array}{ll}
\supportwff\cup\set{\pos A_1,\pos (x_1\ge 1), \pos (x_2\ge 1)},\ & //y=...=\myeuf{11}\\
\supportwff\cup\set{\pos A_1,\pos (x_1\ge 1), \neg (x_2\ge 1)},\  & //y=...=\myeuf{12}\\
\supportwff\cup\set{\pos A_1,\neg (x_1\ge 1), \pos (x_2\ge 2)},\  & //y=...=\myeuf{21}\\
\supportwff\cup\set{\pos A_1,\neg (x_1\ge 1), \neg (x_2\ge 2)},\  & //y=...=\myeuf{22}\\
\supportwff\cup\set{\neg A_1,\pos A_2},& //y=...=\myeuf{3}\\
\supportwff\cup\set{\neg A_1,\neg A_2}& //y=...=\myeuf{4}\\
\end{array}
$

which correspond to the six integrals of Example~\ref{ex:issue1}.
Notice that according to \eqref{eq:wmi_decomposition3} the first four
integrals have to be multiplied by 2, because the partial assignment
\set{A_1} covers two total assignments \set{A_1,A_2} and \set{A_1,\neg A_2}.
Notice also that the disjunction of the six partial assignments, \\
$(A_1\wedge (x_1\ge 1)\wedge (x_2\ge 1)) \vee ... 
\vee (\neg A_1\wedge\neg A_2),
$ matches the definition of \skw{}, which we have computed by progressive enumeration rather
than encoded a priori. \hfill $\diamond$
\end{example}

%\subsection{The \wmipaeuf{} algorithm}
%%%%%%%%%%%%%%%%%%%%%%%%%%%%%%%%%%%%%%%%%%%%%%%%%%%%%%%%%%%%%%%%%
%% The following definitions are to extend the LaTeX algorithmic 
%% package with SWITCH statements and one-line structures.
%% The extension is by 
%%   Prof. Farn Wang 
%%   Dept. of Electrical Engineering, 
%%   National Taiwan University. 
%% 
\newcommand{\SWITCH}[1]{\STATE \textbf{switch} (#1)}
\newcommand{\ENDSWITCH}{\STATE \textbf{end switch}}
\newcommand{\CASE}[1]{\STATE \textbf{case} #1\textbf{:} \begin{ALC@g}}
\newcommand{\ENDCASE}{\end{ALC@g}}
\newcommand{\CASELINE}[1]{\STATE \textbf{case} #1\textbf{:} }
\newcommand{\DEFAULT}{\STATE \textbf{default:} \begin{ALC@g}}
\newcommand{\ENDDEFAULT}{\end{ALC@g}}
\newcommand{\DEFAULTLINE}[1]{\STATE \textbf{default:} }
%% 
%% End of the LaTeX algorithmic package extension.
%%%%%%%%%%%%%%%%%%%%%%%%%%%%%%%%%%%%%%%%%%%%%%%%%%%%%%%%%%%%%%%%%

%%%The novel algorithm is built on top of two key concepts. The main idea is to make \wmipaeuf{} structure-aware by adding information about the atoms of to the problem statements. The conversion algorithm \eufwenc{} discussed previously has this role in the new implementation of \wmipa{}.

%\newpage
\noi
Based on the previous ideas,  we develop \wmipaeuf{},
a novel ``weight structure aware'' variant of \wmipa{}. The pseudocode of \wmipaeuf{} is reported in Algorithm ~\ref{alg:wmipaeuf}.

\noi
As with \wmipa{}, we enumerate the assignments in two main steps: in
the first loop
(rows \ref{line:line2}-\ref{alg:line11}) we generate a set $\MUASTAR{}$ of partial assignments
\muastar{} over the Boolean variables \allA, s.t. \vistarstarmua{} is
\larat-satisfiable and
does not contain Boolean variables anymore. In
Example~\ref{ex:enumeration} 
$\MUASTAR{}\defas\set{\set{A_1},\set{\neg A_1,A_2},\set{\neg A_1,\neg A_2}}$.
In the second loop (rows \ref{line:startPA}-\ref{line:endPA}),
for each \muastar{} in $\MUASTAR{}$ we enumerate the set $\MULARAT$ of
\larat-satisfiable 
partial assignments satisfying \vistarstarmuastar{} 
(that is, on \larat atoms in
$Atoms(\vi\wedge\supportwff)\cup\allpsi$),
we compute the integral \WMINBgen{\mularat}{\wmuastar}{\allx},
multiply it by the $2^{|\allA\setminus\muastar|}$ factor and add
it to the result.
In
Example~\ref{ex:enumeration}, if e.g. $\muastar{}=\set{A_1}$,
$\TA{\predabsg{\vistarstarmuastar}{\atoms{\vistarstarmuastar}}}$
computes the four partial assignments $\{\supportwff\cup\set{(x_1\ge 1),(x_2\ge
    1)},...,\supportwff\cup\set{\neg(x_1\ge 1),\neg (x_2\ge 2)}\}$.

\begin{algorithm}[t]
\caption{{\sf SA-WMI-PA}($\vi$, $\w$, \allx, \allA)
\label{alg:wmipaeuf}}

\begin{algorithmic}[1]
% \STATE return {\sf ComputeVolume}($\vistar$(\allx, $\allA \cup \allB$), $\wstar$)
% \end{algorithmic}
% \end{algorithm}

% \begin{algorithm}[t]
% \caption{{\sf ComputeVolume}($\vi$(\allx, \allA), $\w$)}
% \begin{algorithmic}
\STATE $\MUASTAR\leftarrow\emptyset; vol \leftarrow  0$
\STATE $\newvistar \leftarrow \vi \wedge \supportwff \wedge\eufwenc{}$ \label{line:line1}
\STATE $\MUA \leftarrow\TA{\predabsg{\newvistar}{\allA}}$ \label{line:line2}
\FOR{$\mua \in \MUA$} \label{line:line3start}
    \STATE ${\sf Simplify}(\vistarstarmua)$ \label{alg:simply}
    \IF{\vistarstarmua does not contain Boolean variables}
        \STATE $\MUASTAR \leftarrow \MUASTAR \cup \set{\mua}$ \label{alg:line11}
    \ELSE
    \FOR{$\mua_{residual} \in  \TTA{\predabsg{\vistarstarmua}{\allA}}$} \label{line:line8}
        \STATE $\MUASTAR \leftarrow \MUASTAR \cup \set{\mua \wedge \mua_{residual}}$ \label{line:line9}
    \ENDFOR
    \ENDIF
\ENDFOR \label{line:line3end}
\FOR{$\muastar \in \MUASTAR$} \label{line:startPA}
    \STATE $k \leftarrow |\allA\setminus\muastar|$
    \STATE ${\sf Simplify}(\vistarstarmuastar)$ 
    \IF{${\sf LiteralConjunction}(\vistarstarmuastar)$} \label{line:literalstart}
        \STATE $vol \leftarrow vol + 2^{k}\cdot \WMINBgen{\vistarstarmuastar}{\wmuastar}{\allx}$
        \label{line:literalend}
    \ELSE \label{line:laratstart}
        \STATE $\MULARAT \leftarrow \TA{\predabsg{\vistarstarmuastar}{\atoms{\vistarstarmuastar}}}$ 
        \FOR{$\mularat \in \MULARAT$} 
            \STATE $vol \leftarrow vol + 2^{k} \cdot
            \WMINBgen{\mularat}{\wmuastar}{\allx}$ \label{line:laratend}
        \ENDFOR 
    \ENDIF
\ENDFOR \label{line:endPA}
\STATE return $vol$
\end{algorithmic}
\end{algorithm}

In detail, in row~\ref{line:line1} we extend $\vi\wedge\supportwff$
with $\eufwenc{}$ to provide structure awareness.
(We recall that,
unlike with \wmipa{}, we do not label \larat conditions with fresh
Boolean variables $\allB{}$.)
%
%\noi 
Next, in row~\ref{line:line2} we perform
$\TA{\predabsg{\vistarstarmua}{\allA}}$ to obtain a set $\MUA{}$ of partial
assignments restricted on Boolean atoms \allA. Then, for each
assignment $\mua\in\MUA$ we build the (simplified) residual $\vistarstarmua$.
Since $\mua$ is partial,  $\vistarstarmua$ is not guaranteed to be free of Boolean
variables $\allA$, as shown in Example~\ref{ex:muastar}.
If this is the case, we simply add \mua to $\MUASTAR$, 
otherwise we invoke
$\TTA{\predabsg{\vistarstarmua}{\allA}}$ to assign the remaining
variables and conjoin each 
assignment $\mua_{residual}$ to $\mua$, ensuring that the residual now
contains only \larat{} atoms (rows~\ref{line:line3start}-~\ref{line:line3end}).  
The second loop (rows \ref{line:startPA}-\ref{line:endPA})
resembles the main loop in \wmipa{}, with the only relevant
difference that, since $\muastar$ is partial, the integral is
multiplied by a $2^{|\allA\setminus\muastar|}$ factor.

Notice that in general the assignments \muastar{} are partial even if
the steps in rows \ref{line:line8}-\ref{line:line9} are executed; the set of residual Boolean
variables in $\vistarstarmua$ are a (possibly much smaller) subset of
$\allA\setminus\mua$ because some of them do not occur anymore in
$\vistarstarmua$ after the simplification, as shown in the following example.

\begin{example}
  \label{ex:muastar}
Let $\vi\defas(A_1 \vee A_2 \vee A_3)\wedge(\neg A_1\vee A_2\vee (x \geq
1))\wedge(\neg A_2\vee (x \geq 2))\wedge(\neg A_3\vee (x \leq 3))$,
$\supportwff\defas\inside{x_1}{[0,4]}$
and $\wxa \defas 1.0$.
Suppose \TA{\predabsg{\newvistar}{\allA}} finds the partial
assignment \set{(x \geq 0),(x \leq 4),A_2,(x \geq 1),(x \geq 2),(x
  \leq 3)}, whose projected version is $\mua\defas\set{A_2}$ (row~\ref{line:line2}).
Then $\vistarstarmua$ reduces to $(\neg A_3\vee (x \leq 3))$, so that
$\MUASTAR$ is  \set{\set{A_2,A_3},\set{A_2,\neg A_3}}, avoiding
branching on $A_1$. \hfill $\diamond$
\end{example}
%\APTODO{Questo esempio non e' chiaro, o si leva o si spiega cosa succede dopo..}
%\RSNOTE{io questo esempio lo eliminerei}
%% Sostituito con l'esempio sopra
%\input{example-why-partial-only-partial}

% \RSTODO{evidenzia differenze con \wmipa: partial assignments on
%   $\allA$, partial assignments on $\allpsi$.}

We stress the fact that in our actual implementation, like with that
of \wmipa{}, the potentially-large sets $\MUASTAR$ and $\MULARAT$ are not generated
explicitly. Rather, their elements are
generated, integrated and then dropped one-by-one, so that to avoid
space blowup.

\noi
We highlight two main differences wrt. \wmipa{}.
First, unlike with \wmipa{}, the generated assignments $\mua$ on $\allA$ are partial,
each representing $2^{|\allA\setminus\mua|}$ total ones.
Second, the assignments on (non-Boolean) conditions $\allpsi$ inside
the \mularat{}s are also
partial, whereas with \wmipa{} the assignments to the \allB{}s are total.
This may drastically reduce the number of integrals to compute, as
empirically demonstrated in the next section.

%\newpage

%%%%%%%%%%%%%%%%%%%%%%%%%%%%%%%%%%%%%%%%%%%%%%%%%%%%%%%%%%%%%
%%%
%%%%%%%%%%%%%%%%%%%%%%%%%%%%%%%%%%%%%%%%%%%%%%%%%%%%%%%%%%%%%
\section{Experimental Evaluation}
\label{sec:expeval}

The novel algorithm \wmipaeuf{} is compared to the original \wmipa{} algorithm~\citep{morettin-wmi-aij19}, and the WMI solvers based on KC: XADD~\citep{KolbMSBK18}, XSDD and FXSDD~\citep{kolb2020exploit}. Each of these methods is called from the Python framework pywmi~\citep{pywmi}. For both \wmipa{} and \wmipaeuf{}, we use \mathsat{} for SMT enumeration and \latteintegrale{} for computing integrals. For the KC algorithms we use PSiPSI~\citep{gehr2016psi} as symbolic computer algebra backend. All experiments are performed on an Intel Xeon Gold 6238R @ 2.20GHz 28 Core machine with 128GB of ram and running Ubuntu Linux 20.04. The code of \wmipaeuf{} is freely available at \url{https://github.com/unitn-sml/wmi-pa}. \\
For improved readability, in both the experiments we report runtime using cactus plots, i.e. the single problem instances are increasingly sorted by runtime for each algorithm separately. We highlight how, by construction, problem instances of the same tick of the x-axis are not guaranteed to be the same for different algorithms. Steeper slopes of an algorithm curve means less efficiency.

\begin{figure}[t]
  %\centering \includegraphics[width=1.0\textwidth]{figures/big_synthetic_tree} 
  \centering
   \begin{tabular}{ll}
   \includegraphics[width=0.225\textwidth]{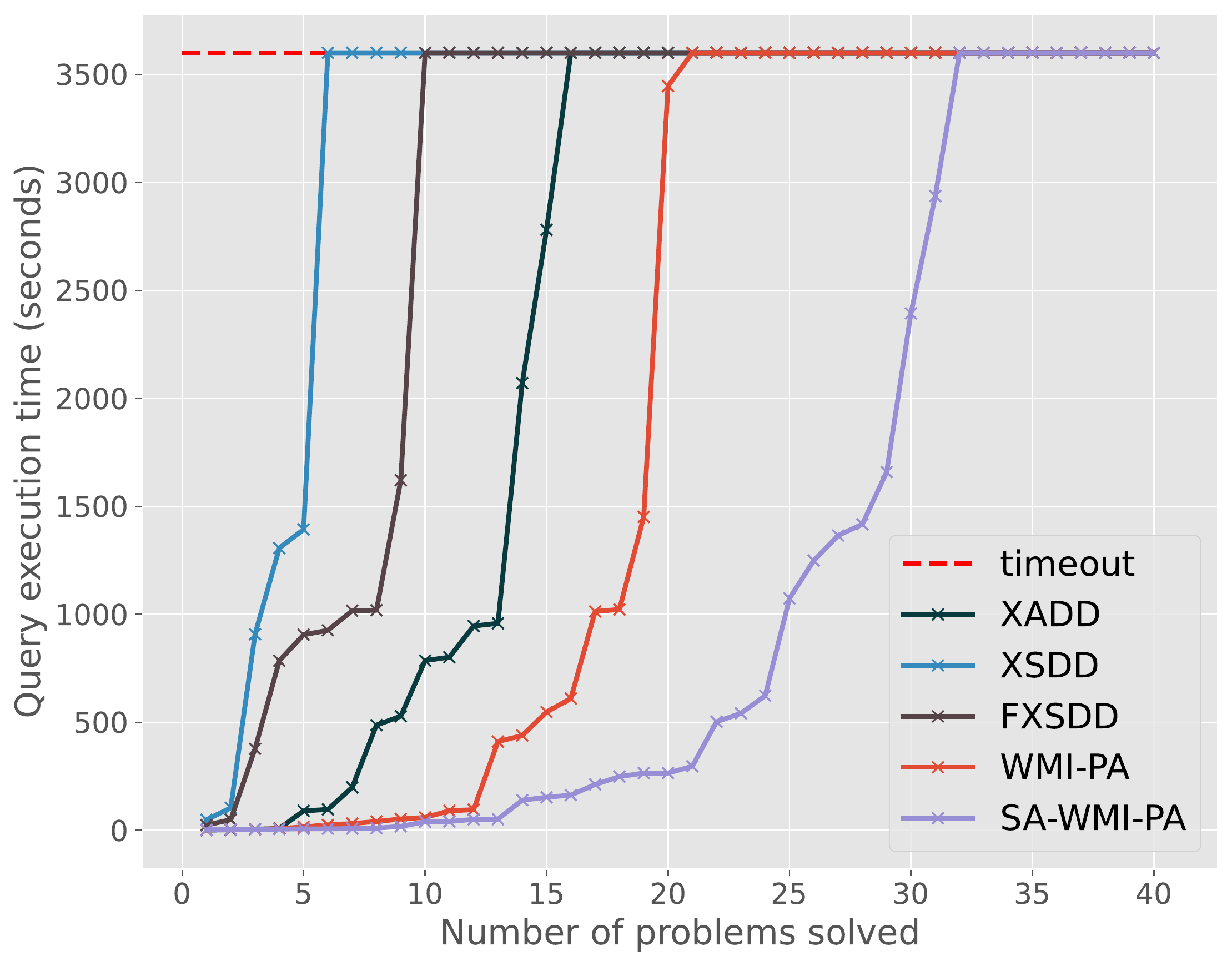}
   &
   \includegraphics[width=0.225\textwidth]{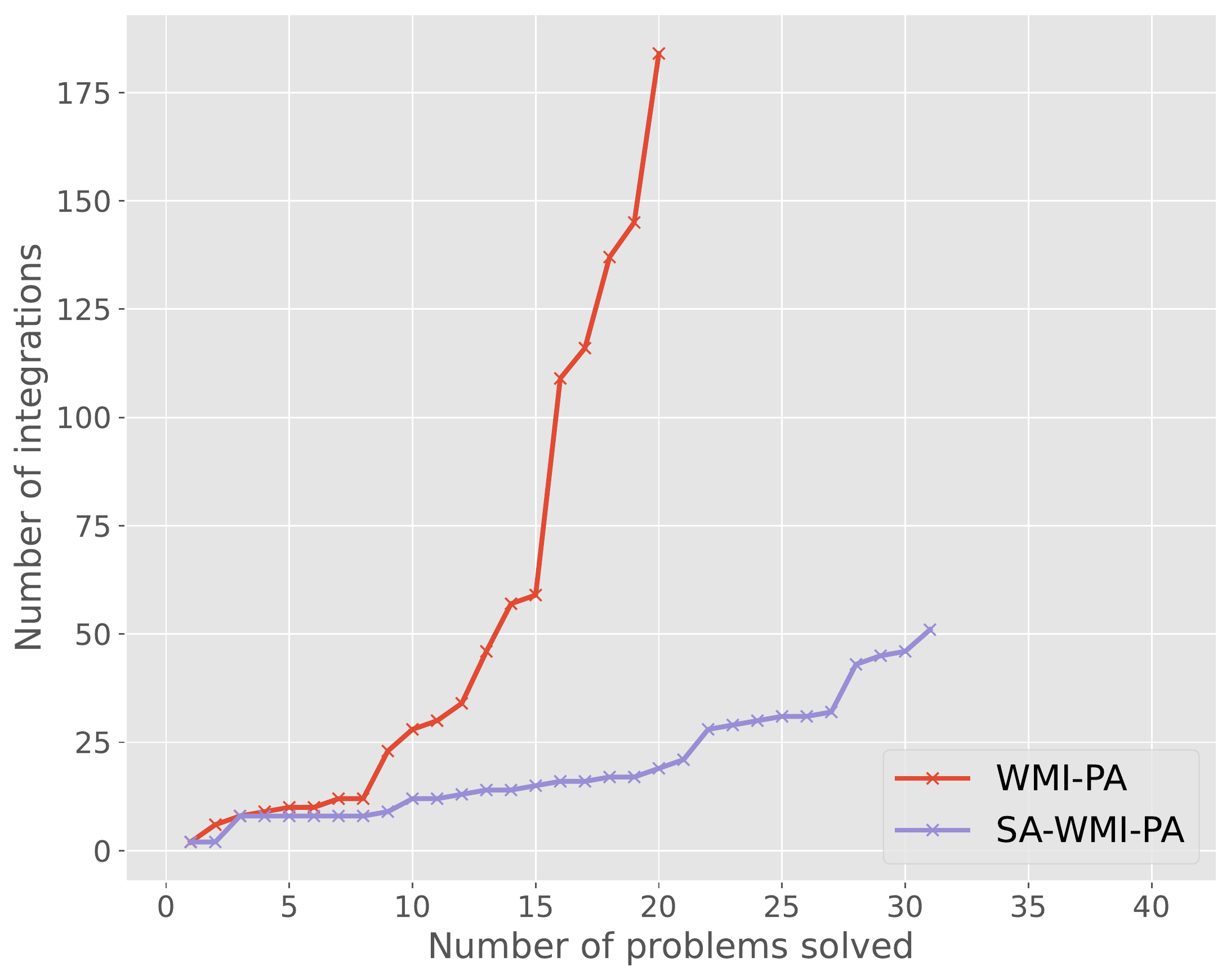}
   \end{tabular}
   \caption{Cactus plots reporting execution time for all methods on the  synthetic
  experiments (left); number of integrals (right) for \wmipa and \wmipaeuf{}.
  \label{fig:synth_tree}}
\end{figure}

\subsection{Synthetic experiments}

%The generation of new synthetic problems have been conceived using the same procedures discussed in \cite{morettin-wmi-aij19}, for both \vixa{} and \wxa{}. To demonstrate the efficiency of \wmipaeuf{} in exploiting information from the structure of \wxa{}, we have focused on testing problems of increasing complexity in respect of the depth of weight function.

%\noi
%The WMI algorithms have been compared according to query execution times; in addition, we provide the number of integrals computed by \wmipa{} and \wmipaeuf{} to demonstrate the potentiality of the structure-aware approach. Results are shown in Figure \ref{fig:synth_tree}  \

%\noi 
%\wmipaeuf{} achieves better performances with respect to both computational times and number of integrals with respect to its predecessor. The differences are more evident with the increasing complexity of the weight function, where the knowledge about its structure can greatly reduce the number of integrals computed by \latteintegrale. The presence of highly structured weight functions justifies the inefficiencies in computation generated by WMI algorithms that exploit symbolic integration.

We first evaluate our algorithm on random formulas and weights, following the experimental protocol of~\cite{morettin-wmi-aij19}. We define two recursive procedures to generate $\larat$ formulae and weight functions with respect to a positive integer number $D$, named depth:

\begin{align} 
\rngTreeB{D} =&
\begin{cases}
\bigoplus_{q=1}^Q \rngTreeB{D-1}\hspace{2.3cm}\text{ if }D > 0 \\
[\neg]\rngAtom \hspace{3.55cm} \text{ otherwise}\\
\end{cases} \nonumber \\
\rngTreeR{D} =&
\begin{cases}
\begin{cases}
\tite{\rngTreeB{D}}{\rngTreeR{D-1}}{\rngTreeR{D-1}} &\\ %&\text{ or}\\
\text{or}\\
\rngTreeR{D-1} \otimes \rngTreeR{D-1} \ \ \ \ \ \ \ \ \ \ \ \text{ if }D > 0\\
\end{cases}
\\
P_{\sf random}(\allx) \hspace{3.2cm} \text{ otherwise}
\end{cases} \nonumber
\label{eq:synth_tree_proc} 
\end{align}

where
$\bigoplus \in \{\bigvee, \bigwedge, \neg \bigvee, \neg \bigwedge\}$,
$\otimes \in \{+,\cdot\}$ and $P_{\sf random}(\allx)$ is a random polynomial function. Using these procedure we generate instances of synthetic problems:
\begin{eqnarray}
%\textstyle
 \chi(\allx,\allA)
 &=& \rngTreeB{D} \wedge \bigwedge_{x \in \allx} \inside{x}{[l_x,u_x]} \nonumber \\
 \wxa
 &=& \rngTreeR{D} \nonumber \\
\viquery(\allx,\allA) &=&  \rngTreeB{D} \nonumber
 %(\bigwedge_{A^\larat_{i,j} \in
 %RandAtoms^\larat_i(\allx)} A^\larat_{i,j}) \vee \bigvee_{A \in
 %R^\Bool_i(\allA)}A \right) \nonumber \\
\label{eq:synth_tree} 
\end{eqnarray}

where $l_x,u_x$ are real numbers such that $\forall x. (l_x < u_x)$. 

\noi
In contrast with the benchmarks used in other recent works~\citep{KolbMSBK18,kolb2020exploit}, the procedure is not strongly biased towards the generation of problems with structural regularities, offering a more neutral perspective on how the different techniques are expected to perform in the wild.
The generated synthetic benchmark contains problems where the number of both Boolean and real variables is set to 3, while the depth of weights functions fits in the range $\lbrack4,7\rbrack$. Timeout was set to 3600 seconds, similarly to what has been done in previous works.

\noi
\ignoreinshort{
Results in this paper are shown in the following way. For each WMI algorithm, instances that provide the result within the timeout are ranked with respect to query execution times (and additionally, for PA-based methods, the number of computed integrals) and plotted in ascending order. We highlight how, by construction, problem instances of the same tick of the x-axis are not guaranteed to be the same for different algorithms. The number of instances on the x-axis hence can be interpreted as the number of problems the considered algorithm solved within timeout. Steeper slopes of an algorithm curve means less efficiency. \textcolor{red}{\textbf{Paolo}: questo paragrafo lo riscriverei sopra, inoltre "number of instances on the x-axis hence can be interpreted as the number of problems the considered algorithm solved within timeout" mi sembra ambiguo.}
\textcolor{blue}{\textbf{Giuseppe}: puó andare bene muovere il paragrafo ad intro degli esperimenti. Se la frase sembra ambigua, si potrebbe rimuovere e mantenere la questione della slope e la sua interpretazione, che é il cuore dei cactus.}
}

\noi
In this settings, the approaches based on a SMT oracle clearly outperform those based on KC (Fig. \ref{fig:synth_tree} left). In addition, \wmipaeuf{} greatly improves over \wmipa{}, thanks to a drastic reduction in the number of integrals computed (Fig. \ref{fig:synth_tree} right), with the advantage of our approach getting more evident when the weight functions are deeper.

\begin{figure}[t]
  %\centering \includegraphics[width=1.0\textwidth]{figures/big_synthetic_tree} 
  \centering
   \begin{tabular}{cc}
   %\includegraphics[width=0.175  \textwidth]{plots/MLC/time_uai_dets-100-200-5-0.0_cactus.pdf}
   %&
   \includegraphics[width=0.225\textwidth]{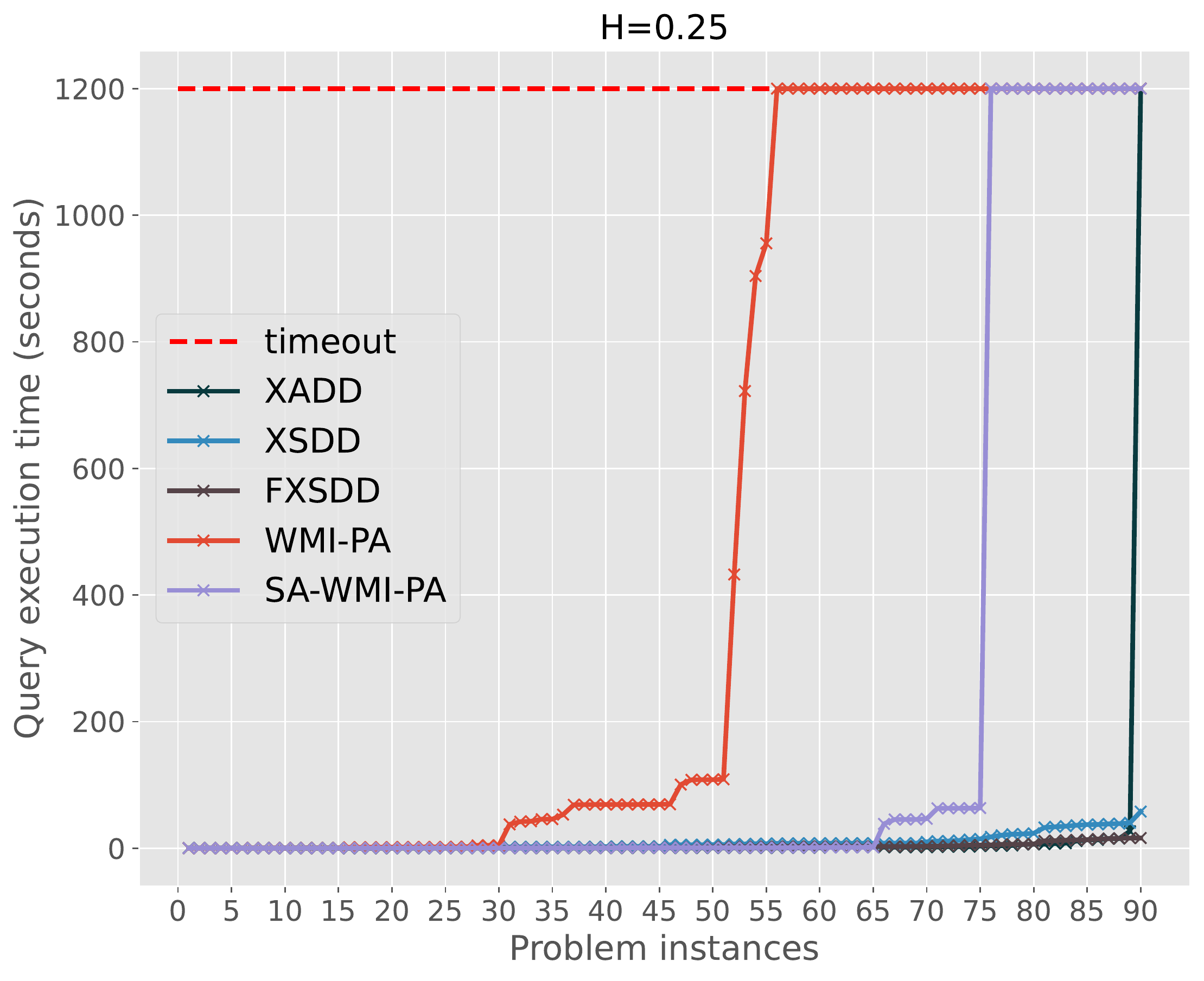}
   &
   \includegraphics[width=0.225\textwidth]{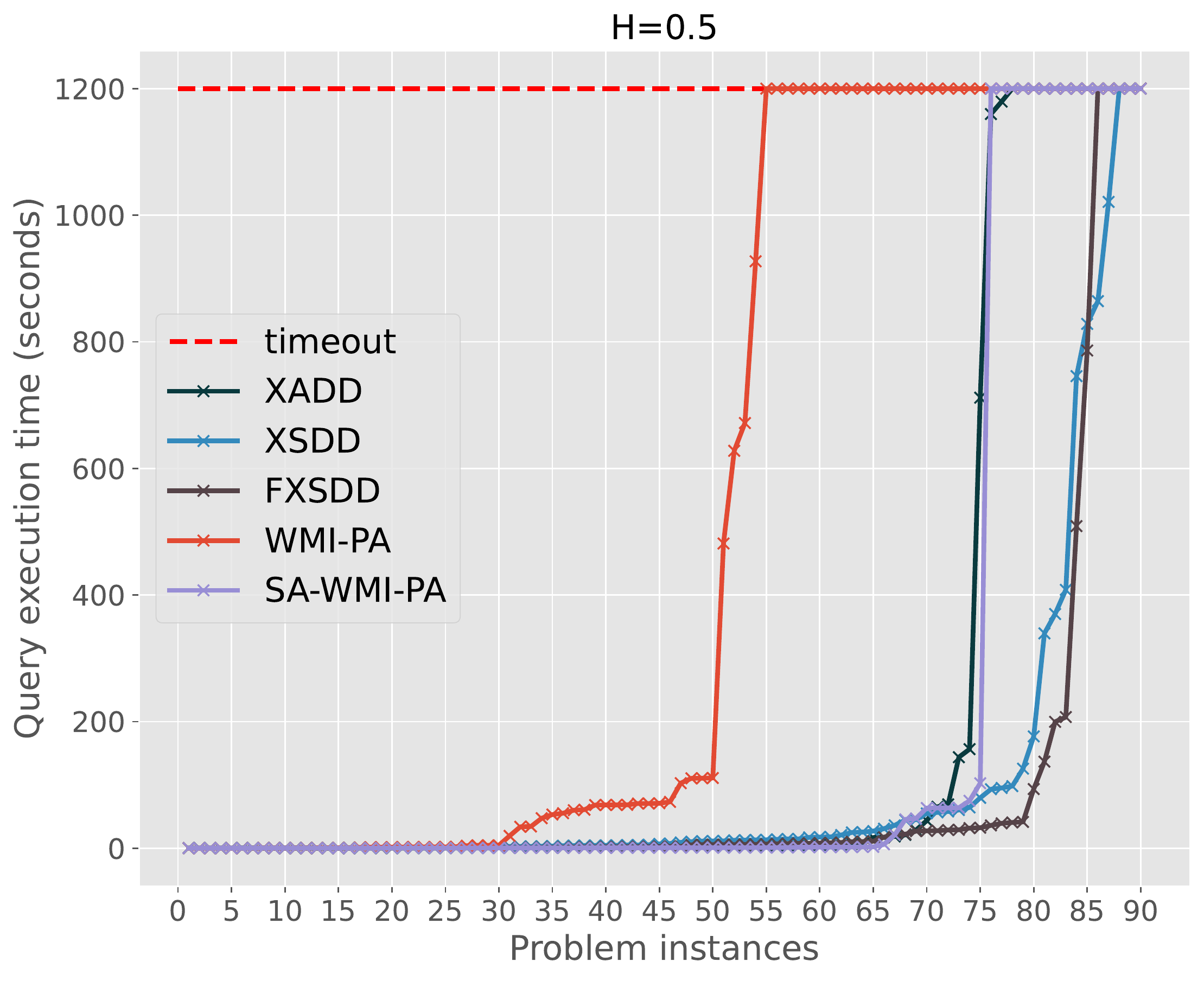}
   \\
   \includegraphics[width=0.225\textwidth]{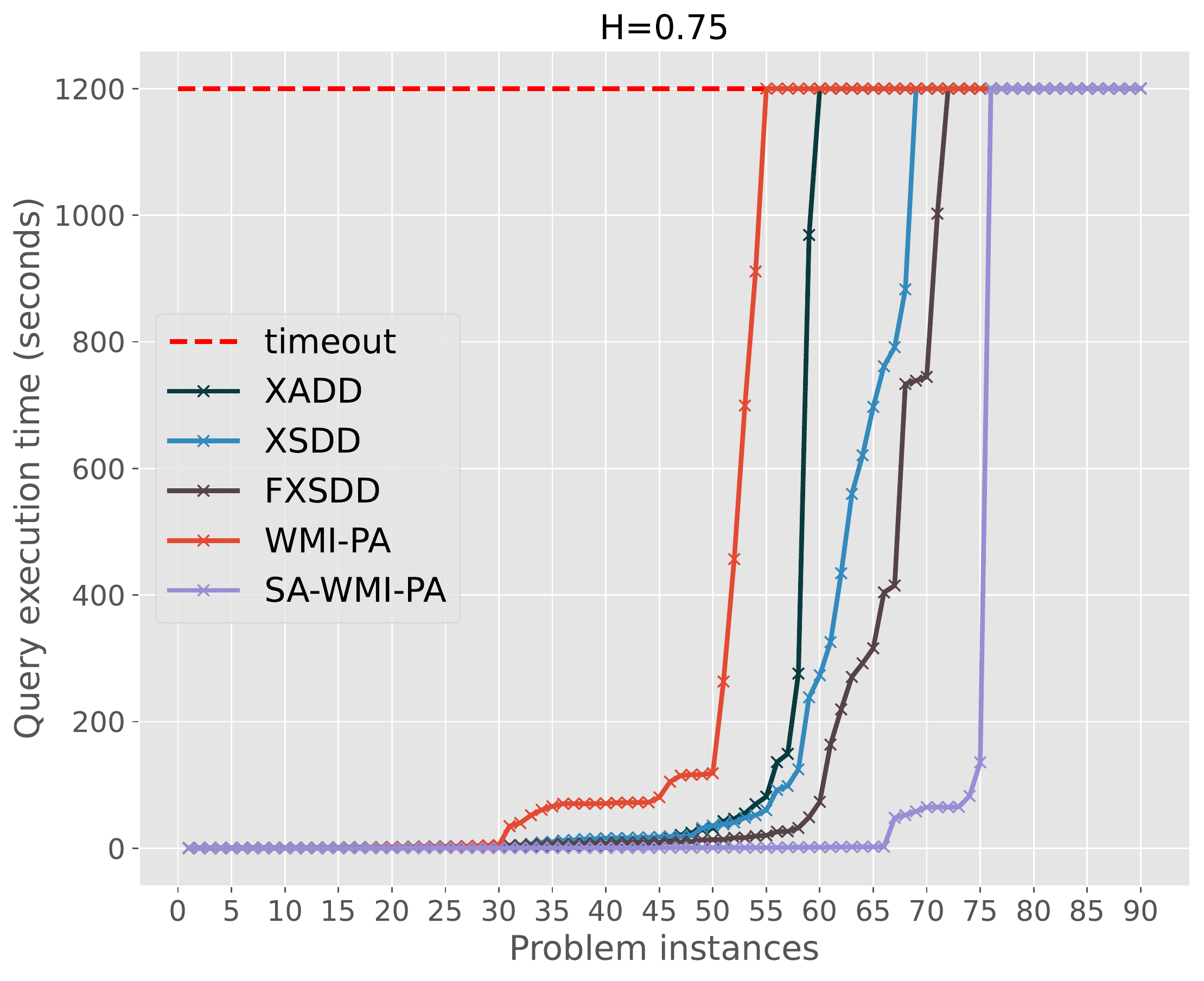}
   &
   \includegraphics[width=0.225\textwidth]{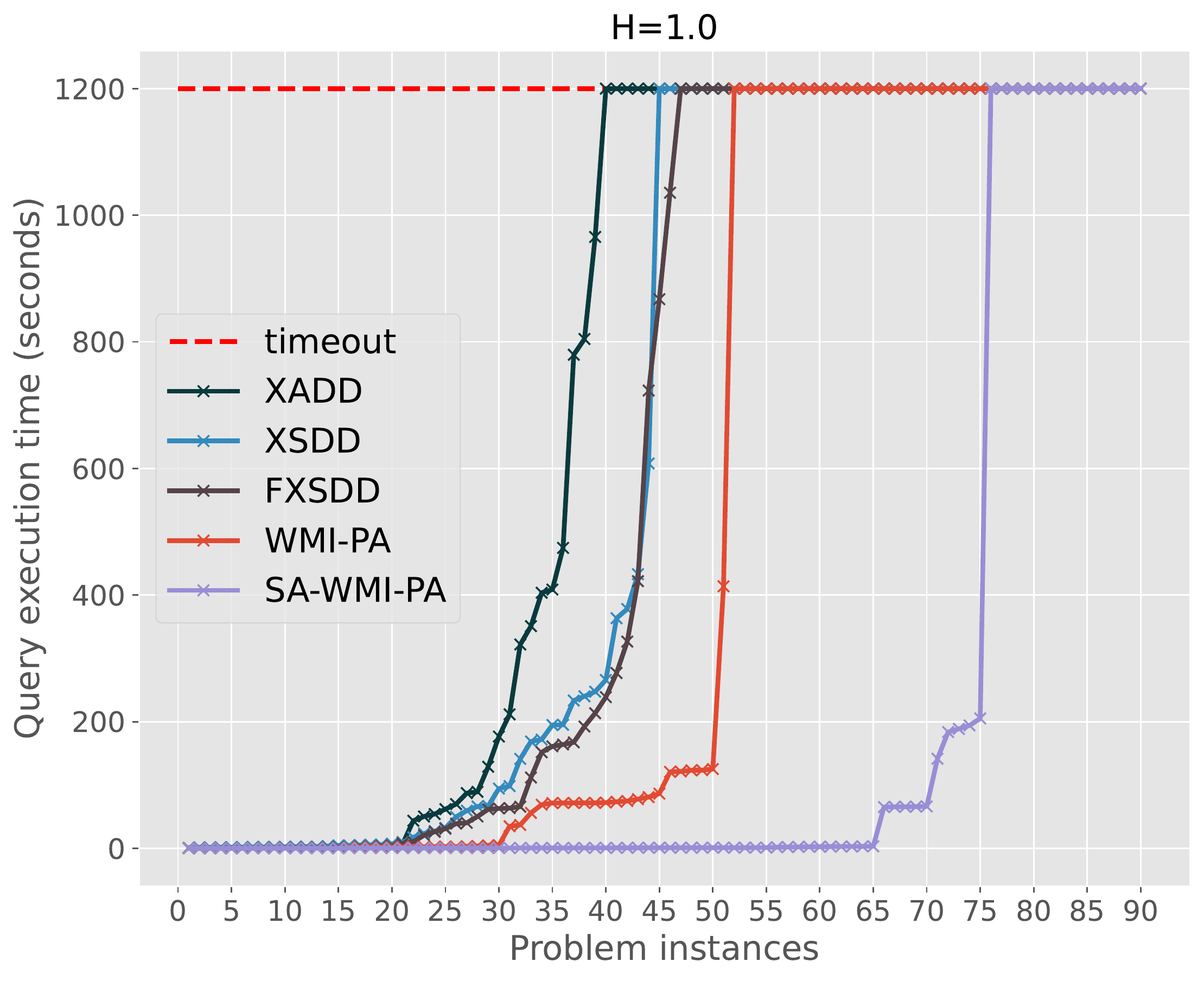}
   \end{tabular}
   \caption{Cactus plots representing average query execution times and standard deviation in seconds on a set of DET problems with $H \in \{0.25, 0.5, 0.75, 1\}$.
  \label{fig:detplot_time}}
\end{figure}

\subsection{Density Estimation Trees}

We explore the use of WMI solvers for marginal inference in real-world probabilistic models. In particular, we considered \emph{Density Estimation Trees} (DETs)~\citep{ram2011density}, hybrid density estimators encoding piecewise constant distributions. Having only univariate conditions in the internal nodes, DETs natively support tractable inference when single variables are considered. Answering queries like $Pr(X \le Y)$ requires instead marginalizing over an oblique constraint, which is reduced to WMI:
$Pr(X \le Y) = \frac{WMI((X \le Y) \wedge \chi_{DET}, w_{DET})}{WMI( \chi_{DET}, w_{DET})}$.
Being able to address this type of queries is crucial to apply WMI-based inference to e.g. probabilistic formal verification tasks, involving constraints that the system should satisfy with high probability.
%WMI-based inference not only can broaden the range of queries that can be answered, but is crucial for quantitative formal verification tasks, using SMT(\larat) as a flexible language for expressing properties or constraints that the system should satisfy with high probability.
%

\noi
We considered a selection of hybrid datasets from the UCI repository~\citep{Dua2019uci}, reported in Table 1 in the Appendix. \ignoreinshort{\textcolor{red}{together with statistics on the runtime aggregated by dataset}.} Following the approach of~\cite{morettin2020learning}, discrete numerical features were relaxed into continuous variables, while $n$-ary categorical features are one-hot encoded with $n$ binary variables.

After learning a DET on each dataset, we generated a benchmark of increasingly complex queries, 5 for each dataset, involving a ratio $H \in [0,1]$ of the continuous variables. More specifically, the queries are linear inequalities involving a number of variables $k = max(1, \lfloor H \cdot |\allx|\rfloor)$.
Figure~\ref{fig:detplot_time} depicts the runtime of the algorithms for $H \in \{0.25, 0.5, 0.75, 1\}$. Timeout was set to 1200 seconds.
 KC approaches have an edge for the simplest cases ($H \le 0.5$) in which substantial factorization of the integrals is possible. 
Contrarily to many other probabilistic models, which are akin to the case in Section~\ref{sec:dd_issues}, DETs are well-suited for KC-based inference, due to the absence of arithmetic operations in the internal nodes.
When the coupling between variables increases, however, the advantage of decomposing the integrals is overweight by the combinatorial reasoning capabilities of \wmipaeuf{}. We remark that
\wmipaeuf{} is agnostic of the underlying integration procedure, and thus in principle it could also incorporate a symbolic integration component. \ignoreinshort{This promising direction is left for future work.} 
% \begin{figure}
%     \centering
%     \includegraphics[width=0.5\textwidth]{plots/MLC/time_uai_dets-100-200-666_cactus.pdf}
%     \caption{Cactus plot representing average query execution times and standard deviation in seconds for a set of DET problems.}
%     \label{fig:detplot}
% \end{figure}
%\begin{figure*}[t]
  %\centering \includegraphics[width=1.0\textwidth]{figures/big_synthetic_tree} 
%  \centering
%   \begin{tabular}{ccc}
%   \includegraphics[width=0.31\textwidth]{plots/MLC/n_integrations_uai_dets-100-200-5-0.0_cactus.pdf}
%   &
%   \includegraphics[width=0.31\textwidth]{plots/MLC/n_integrations_uai_dets-100-200-5-0.5_cactus.pdf}
%   &
%   \includegraphics[width=0.31\textwidth]{plots/MLC/n_integrations_uai_dets-100-200-5-1.0_cactus.pdf}
%   \end{tabular}
%   \caption{Cactus plot representing average number of integrals and standard deviation for \wmipa and \wmipaeuf{} on a set of DET problems with different qhardness: $0.0$ (left), $0.5$ (center), $1.0$ right.
%  \label{fig:detplot_int}}
%\end{figure*}
%%%%%%%%%%%%%%%%%%%%%%%%%%%%%%%%%%%%%%%%%%%%%%%%%%%%%%%%%%%%%
%%%
%%%%%%%%%%%%%%%%%%%%%%%%%%%%%%%%%%%%%%%%%%%%%%%%%%%%%%%%%%%%%
%\section{Conclusions and Future Developments}
\section{Conclusion}
\label{sec:concl}

% The difficulty of dealing with densely coupled problems has been a
% major obstacle to a wider adoption of hybrid probabilistic inference
% technology beyond academia. In this paper we made a significant step
% towards removing this obstacle. Starting from the identification of
% the limitations of existing solutions for WMI, we proposed a novel
% algorithm combining predicate abstraction with weight-structure
% awareness, thus unleasing the full potential of SMT-based
% technology. Despite the substantial gains of the SA-WMI-PA algorithm,
% there is still room for improvement. As the original WMI-PA, SA-WMI-PA
% is agnostic of the underlying integration procedure. Combining
% strucuture-awareness with symbolic integration capabilities could
% boost the performance of the algorithm in the cases of partial
% coupling when knoweldge compilation technology is still competitive. 
% \APTODO{check, mention other future work?}

We presented the first SMT-based algorithm for WMI that is aware of
the structure of the weight function. This is particularly beneficial
when the piecewise density defined on top of SMT($\larat$) constraints
is deep, as it is often the case with densities learned from data.
We evaluated our algorithmic ideas on both synthetic and real-world
problems, obtaining state-of-the-art results in many settings of
practical interest.
Providing unprecedented scalability in the evaluation of complex
probabilistic SMT problems, this contribution directly impacts the use
of WMI for the probabilistic verification of systems.
While the improvements described in this work drastically reduce the
number of integrations required to compute a weighted model integral,
the integration itself remains an obstacle to scalability. Combining
the fast combinatorial reasoning offered by SMT solvers with symbolic
integration is a promising research direction.

%%%%%%%%%%%%%%%%%%%%%%%%%%%%%%%%%%%%%%%%%%%%%%%%%%%%%%%%%%%%%
%%%
%%%%%%%%%%%%%%%%%%%%%%%%%%%%%%%%%%%%%%%%%%%%%%%%%%%%%%%%%%%%%

\begin{acknowledgements}
This research was partially supported by TAILOR, a project
funded by EU Horizon 2020 research and innovation programme under GA No 952215 and from the European Research Council (ERC) under the European Union’s Horizon
2020 research and innovation programme (grant agreement
No. [694980] SYNTH: Synthesising Inductive Data Models). The authors would like to thank Pedro Zuidberg Dos Martires for his help with XSDDs and FXSDDs. \end{acknowledgements}

%\FloatBarrier
%% The file named.bst is a bibliography style file for BibTeX 0.99c
%\bibliographystyle{named}
\bibliography{spallitta_478}
%\bibliography{ijcai22}

\clearpage
% \section*{Appendix: Proofs of the Theorems}
%% there are no "theorems"

\end{document}

\maketitle

\section*{\eufwenc{} definition}

Algorithm~\ref{algo:convert} is such that,
given a \FIUC{} weight function \w{} on conditions
$\allpsi$, \convert{}($\w$,$\emptyset{}$) returns 
  \tuple{\w',\defs,\newvars} s.t. 
$\eufwenc{}\defas(y=\w')\wedge\bigwedge_{\vi_i\in\defs}\vi_i$, on
variables $\allx\cup\ally,\allA$.

\begin{algorithm}[h]
  \caption{
    \convert{}($\term$, $\conds{}$)\\
    returns \tuple{\w',\defs,\newvars} \\
    \w': the term \w{} is rewritten into\\
    \conds{}: the current partial assignment to conditions \allpsi, \\representing
    the set of conditions which $\w$ depends on\\
    \defs{}: a set of definitions in the form $y_i=\w_i$ needed to
    rewrite \w{} into \w{}'\\
    \ally{}: newly-introduced variables labeling if-then-else terms\\
  % \convert{}($\term$, $\conds{}$) returns
  % \tuple{\term',\defs,\newvars} s.t.\\
  % $(\conds \imp \exists y. y=\term)
  % \Longleftrightarrow
  % \exists \newvars.(\defs \wedge (\conds \imp \exists y. y=\term'))$\\
  $f^g$: uninterpreted function naming the
  function/operator $g$
%  \TODO{riscrivere come case-of...?}
%  \TODO{mettere \defs{} come variabile globale/static?}
%  \TODO{RS: togliere mutex?}
}
\begin{algorithmic}[1]
%  \IF{(\{$\term$ contains no conditions\})}
  \IF{(\{$\term$ constant or variable\})}
  \STATE {\bf return} \tuple{\term,\emptyset{},\emptyset{}}
  \ENDIF
  \IF{($\term==(\term_1 \bowtie \term_2)$, $\bowtie\ \in\set{+,-,\cdot,/}$)}
  % \STATE $\tuple{\term_1',\defs_1}=\convert{}(\term_1,\conds{})$
  % \STATE $\tuple{\term_2',\defs_2}=\convert{}(\term_2,\conds{})$
  \STATE $\tuple{\term_i',\defs_i,\newvars_i}=\convert{}(\term_i,\conds{})$, $i\in{\{1,2\}}$
  \STATE {\bf return} \tuple{f^{\bowtie}(\term_1',\term_2'),\defs_1\cup \defs_2,\newvars_1\cup \newvars_2}
  \ENDIF
  \IF{($\term==g(\term_1,...,\term_k)$, $g$ unconditioned)}
  \STATE $\tuple{\term_i',\defs_i,\newvars_i}=\convert{}(\term_i,\conds{})$, $i\in{1,...,k}$
  \STATE {\bf return} \tuple{f^g(\term_1',...,\term_k'),\cup_{i=1}^k \defs_i,\cup_{i=1}^k \newvars_i}
  \ENDIF
  \IF{($\term==({\sf If}\ \psi\ {\sf Then}\ \term_{1}\ {\sf Else}\ {\term_{2}})$)}
  \STATE $\tuple{\term_1',\defs_1,\newvars_1}=\convert{}(\term_1,\conds{}\cup\set{\pos\psi})$
  \STATE
  $\tuple{\term_2',\defs_2,\newvars_2}=\convert{}(\term_2,\conds{}\cup\set{\neg\psi})$
  \STATE {\bf let} $y$ be a fresh variable
  \STATE $\defs=\defs_1\cup\defs_2\ \cup$ %\Comment{y fresh}
  \STATE \ \ \ \ $(\bigvee_{\psi_i\in\conds}\neg\psi_i\vee \neg \psi\vee (y=\term_1')\ \cup$
  \STATE \ \ \ \ $(\bigvee_{\psi_i\in\conds}\neg\psi_i\vee\pos \psi\vee (y=\term_2')\ \cup$
  \STATE \ \ \ \ $(\bigvee_{\psi_i\in\conds}\neg\psi_i\vee\neg(y=\term_1')\vee\neg(y=\term_2'))$
\STATE $\newvars = \newvars_1\cup\newvars_2\cup\set{y}$ 
  \STATE {\bf return} \tuple{y,\defs,\newvars}
  \ENDIF
\end{algorithmic}
\label{algo:convert}
%\TODO{RS: Check: nella riga 2 \emptyset{} e' corretto? }

\end{algorithm}

\begin{table}[tbh!]
    \centering
    \begin{tabular}{|l|c|c|c|c|}
\hline
Dataset & $|\allA|$ & $|\allx|$ & \# Train & \# Valid\\
\hline
balance-scale & 3 & 4 & 1875 & 205 \\
iris & 3 & 4 & 450 & 50 \\
cars & 33 & 7 & 2115 & 234 \\
diabetes & 1 & 8 & 4149 & 459 \\
breast-cancer & 12 & 4 & 1650 & 180 \\
glass2 & 1 & 9 & 970 & 100 \\
glass & 7 & 9 & 1280 & 140 \\
breast & 1 & 10 & 4521 & 495 \\
solar & 25 & 3 & 2522 & 273 \\
cleve & 17 & 6 & 2492 & 266 \\
hepatitis & 14 & 6 & 940 & 100 \\
heart & 3 & 11 & 2268 & 252 \\
australian & 34 & 6 & 6210 & 690 \\
crx & 38 & 6 & 6688 & 736 \\
german & 41 & 10 & 12600 & 1386 \\
german-org & 13 & 12 & 15000 & 1650 \\
auto & 56 & 16 & 2522 & 260 \\
anneal-U & 74 & 9 & 21021 & 2301 \\

\hline
\end{tabular}
\caption{UCI datasets considered in our experiments. We report the number of Boolean and continuous variables, training and validation data size.}
    \label{tab:datasets}
\end{table}